%% file: main.tex
\documentclass[twoside]{article}
\usepackage[accepted]{aistats2023}

\usepackage[round]{natbib}

\input{preamble}

\begin{document}

\runningtitle{The Role of Codeword-to-Class Assignments in Error Correcting Codes:
 An Empirical Study}

\twocolumn[

\aistatstitle{The Role of Codeword-to-Class Assignments in Error-Correcting Codes:
 \protect\\
 An Empirical Study}

\aistatsauthor{
Itay Evron$^*$
\And Ophir Onn$^*$
\And Tamar Weiss Orzech \And Hai Azeroual \And Daniel Soudry
%\thanks{$^*$ Equal Contribution.}
}

\aistatsaddress{
Department of Electrical and Computer Engineering,
Technion, Israel}]

\begin{abstract}
Error-correcting codes (ECC) are used to reduce multiclass classification tasks to multiple binary classification subproblems.
\linebreak
In ECC, classes are represented by the rows of a binary matrix, corresponding to codewords in a codebook.
Codebooks are commonly either predefined \emph{or} problem-dependent.
Given predefined codebooks, 
codeword-to-class assignments are traditionally overlooked, and codewords are implicitly assigned to classes arbitrarily.
\linebreak
Our paper shows that these assignments play a major role in the performance of ECC.
Specifically, we examine similarity-preserving assignments, where similar codewords are assigned to similar classes.
Addressing a controversy in existing literature,
our extensive experiments
confirm that similarity-preserving assignments induce easier subproblems
and are superior to other assignment policies in terms of their generalization performance.
We find that similarity-preserving assignments make
\emph{predefined} codebooks 
become problem-dependent, 
without altering other favorable codebook properties.
Finally, we show that our findings
can improve predefined codebooks dedicated to extreme classification.
\end{abstract}

\Section{Introduction}

Error-correcting codes (ECC) have been long used in machine learning as a reduction from multiclass classification tasks to binary classification tasks \citep{dietterich1994solving}.
This scheme encodes classes using rows of a binary matrix
called a codebook.
The codebook columns induce binary partitions of classes, or subproblems, to be learned using any binary classification algorithm.

Recently, error-correcting codes have been used as output embeddings of deep networks 
\citep{yang2015deepECOC, rodriguez2018beyond,kusupati2021llc},
on top of features extracted by deep CNNs \citep{Dorj2018ECOCSVM},
and as a means to combine ensembles of several networks \citep{zheng2018ensemble}.
Moreover, they were recently used for their robustness
in adversarial learning \citep{verma2019adversarial,gupta2020integer,song2021robust}
and for their redundancy in regression tasks
\citep{shah2022regression}
and heterogeneous domain adaptation \citep{zhou2019domain_adapation}.

In extreme multiclass classification, 
where the number of classes is extremely large, 
ECC can be particularly beneficial.
Several works \citep{jasinska2016ltls, evron2018wltls}
employed ECC to shrink the output space, decreasing the number of learned predictors, as well as the prediction time, to
\emph{logarithmic} in the number of classes.
In comparison, both one-hot encoding and hierarchical models train a linear number of predictors 
(even though the latter enjoy a logarithmic prediction time).

The first step in employing ECC consists of selecting a \emph{good} codebook.
Some codebook properties are universally important for error correction, \eg  the minimum hamming distance between rows.
Other properties are only important in some regimes, \eg the decoding complexity which is essential mainly in extreme classification.

Roughly, codebooks can be divided into two categories: \emph{predefined codebooks} and \emph{problem-dependent codebooks}.
Predefined codebooks are independent of the problem at hand, 
but offer 
simplicity (\eg random codebooks),
favorable error-correction properties 
(\eg Hadamard codebooks in \citealp{zhang2003hadamard}
or optimized codebooks in \citealp{gupta2022ECOCcoloring}),
or
regime-specific advantages like fast decoding algorithms \citep{evron2018wltls}.
On the other hand, problem-dependent approaches attempt to induce binary subproblems that are tailored for a given dataset, often by balancing against other codebook properties.

Problem-dependent codebooks are commonly designed by optimizing over codebooks while taking 
\emph{class-similarity} into account.
However, there are two \emph{opposite} intuitions in the literature as to \emph{how} to incorporate class-similarity in the design process.
Some works follow an intuition that to induce easy subproblems,
similar classes should be encoded by \emph{similar} codewords 
\citep{zhang2009spectral, 
cisse2012compact,
zhao2013sparse,
zhou2016confusion,
rodriguez2018beyond}. 
In contrast, other works encode similar classes by \emph{distant} codewords to improve the error correction between hardly-separable classes
\citep{pujol2008incremental,martin2017factorization,youn2021construction,gupta2020integer}.
\linebreak
We examine this \emph{controversy} in depth
and provide evidence from multiple regimes 
that generalization is superior when
encoding
similar classes by \emph{similar} codewords.

In \emph{predefined} codebooks, 
the mapping between codewords and classes, \ie the \emph{codeword-to-class assignment}, is usually set arbitrarily (\eg using a random assignment).
\citet{dietterich1994solving} showed that randomly-sampled assignments perform similarly, and since, these assignments have been commonly overlooked.

Our paper shows that \emph{codeword-to-class assignments \linebreak do matter} and cause a large variation in the performance of many predefined codebooks (\secref{sec:variability}).
We explain this by showing that, given a codebook, some assignments induce substantially easier binary subproblems than other assignments do (\secref{sec:exp_train_loss}).
Moreover, we show that the easiest subproblems are induced by assigning similar codewords to similar classes (\secref{sec:exp_class_similarity}).

Finally, we employ our observations on extreme multiclass classification datasets (having 1K to 104K classes).
By assigning similar codewords to similar classes, we significantly improve predefined extreme classification codebooks that enjoy fast decoding algorithms  (\secref{sec:extreme_experiments}).  

To the best of our knowledge, 
this is the first work to point out the large performance variation 
explained solely by codeword-to-class assignments,
and to explicitly examine
these assignments as a means to control the difficulty of the induced learning-subproblems
in problem-\emph{independent}
predefined codebooks.
We conclude that choosing an informed assignment improves predefined codebooks
by turning them problem-\emph{dependent} 
and better suited for the solved task.
Importantly, other useful properties of these codebooks are \emph{not} harmed in this process.

\Section{Error-correcting codes (ECC)}
\label{sec:ecoc}

Error-correcting codes are widely used for transmitting messages over noisy channels in communication systems, storage systems, and more.
By adding redundant bits to transmitted messages, 
the receiver can recover messages despite errors caused by a disruptive channel \citep{roth2006coding}.

\paragraph{Training.}
The seminal work of \citet{dietterich1994solving}
employed error-correcting codes to encode the
$K$ classes of a classification dataset.
They set a binary codebook 
$\M\in\left\{-1, +1\right\}^{K\times \ell}$
with $K\in\mathbb{N}$ codewords (each belonging to one class)
and $\ell$ columns 
(where $\ell\ge\log_2 K$).
Each column induces a \emph{binary subproblem}, \ie a binary partition of classes.
Each such subproblem is learned using a base learner $\mathcal{A}$
%: \left(\mathcal{X} \times \left\{-1,+1\right\}\right)^m$ 
(\ie a binary classification learning algorithm),
yielding $\ell$ predictors $f_1,\smalldots,f_\ell:
\mathcal{X}\to\doubleR$.
More formally,
given a training set $\left\{  \left(\vx_i,y_i\right) \right\}_{i=1}^{m}$,
where $x_i\in\mathcal{X}$ and $y_i \in \range{K} \triangleq \left\{1, \smalldots, K\right\}$,
the~$j$th~predictor is the output of $\mathcal{A}$ when trained using the induced binary labels $M_{y_i, j}$:
\begin{align}
\label{eq:train_predictor}
    f_j =
    \mathcal{A}\left(
        \left\{
            \left(\vx_i, M_{y_i, j}\right)
        \right\}_{i=1}^{m}
    \right)~.
\end{align}

\paragraph{Prediction.}
At prediction time, an example $\vx\!\in\!\mathcal{X}$
is treated as a transmitted message encoding the unknown class 
$y\!\in\! \range{K}$.
The $\ell$ predictors' scores for $\vx$
constitute the vector 
$\vf\!\left(\vx\right)
\triangleq
\left[f_1\!\left(\vx\right)\!, 
\smalldots,
f_\ell\!\left(\vx\right)\right]^\top
\!$.  %
These scores can be prediction margins from a linear model, 
confidences from a probabilistic model, outputs of a neural network, etc.

Finally, the prediction vector $\vf\!\left(\vx\right)$ is \emph{decoded} into a codeword belonging to a class.
The simplest approach is 
\emph{hard decoding} that consists of finding the nearest neighbor, that is, the codeword closest (in Hamming distance) to the thresholded prediction vector, $\sgn\!\left(\vf\!\left(\vx\right)\right)\in\left\{-1, +1\right\}^{\ell}$. 

Hard decoding ignores the score magnitudes which entail valuable information for prediction. 
As a remedy, \emph{soft decoding}, or loss-based decoding \citep{allwein2000loss}, minimizes a decoding loss $\mathcal{L}:\doubleR\to\doubleR_{\ge 0}$:
\begin{align}
\label{eq:soft-decoding}
    \hat{y}\left(\vx\right)=
    \argmin_{y\in\range{K}}
    \tsum_{j=1}^{\ell}
    \mathcal{L}
    \left(
        M_{y,j} f_j\left(\vx\right)
    \right)~.
\end{align}
Two popular decoding losses are the hinge loss ${\mathcal{L}\left( z \right)=\max\left\{ 0, 1-z\right\}}$
and the exponential loss
${\mathcal{L}\left( z \right)=e^{-z}}$.
Notice that soft decoding generalizes hard decoding with 
${\mathcal{L}\left( z \right)
=
\frac{1-\sgn\left(z \right)}{2}}$.

We illustrate the entire ECC scheme in \appref{app:basics}.

\paragraph{Multiclass error upper bound.}
\citet{allwein2000loss} proved an insightful upper bound%
\footnote{
\citet{zhou2019nary}
derived a  bound for more general N-ary codes
(where subproblems are also multiclass instead of binary), 
but this remains out of our scope in this work.}
that will facilitate our discussion throughout this paper.
Let 
\begin{align}
    \varepsilon
\triangleq
    \varepsilon(\M, \mathcal{L})
=
\frac{1}{m\ell}
\tsum_{i=1}^{m}
\tsum_{j=1}^{\ell}
    \mathcal{L}
    \left(
        M_{y_i,j} f_j\left(\vx_i\right)
    \right)
\label{eq:average_binary_loss}
\end{align}
be the \emph{average binary loss} of the binary predictors
on a given training set $\left\{
\left(\vx_i,y_i\right)
\right\}_{i=1}^{m}$
with respect to a codebook $\M$
and a decoding loss $\mathcal{L}$.
Assume %that
$\mathcal{L}$ satisfies mild conditions (\eg convexity is sufficient).
Then, the \emph{multiclass training error} 
when decoding with $\mathcal{L}$ is upper bounded as:
\begin{align}
    \frac{1}{m}
    \tsum_{i=1}^{m}
    \mathbb{I}[y_i \neq \hat{y}(\vx_i)]
    \le
    \frac{\ell \varepsilon}{\rho \mathcal{L}\left(0\right)}~,
    \label{eq:bound}
\end{align}
where 
%the~$k$th~row~of $\M$ is denoted by $\M_{k,:}$,
${\rho
\triangleq
\rho(\M)=
\min_{a\neq b} 
\frac{1}{2}
\norm{\M_{a,:}\!-\M_{b,:}}_1}$
is
the codebook's \emph{minimum inter-row Hamming distance}
($\M_{a,:}$ being the~$a$th~row~of $\M$)
%
% ${\rho\triangleq
% \min_{a\neq b} 
% \frac{1}{2}
% \tnorm{\underbrace{\M_{a,:}}_{a\text{th row}}-\M_{b,:}}_1}$,
%
and ${\mathcal{L}\left(0\right)}$ is a scaling factor of $\mathcal{L}$.
%

%\pagebreak

\pagebreak
        
\subsection{Properties of a Good Codebook}
\label{sec:codebook_properties}
We now 
review favorable properties of error-correcting codebooks.
The first two properties are discussed more often in the literature (\eg \citealp{dietterich1994solving,zhang2003hadamard}),
while the latter two are seldom addressed despite their importance.
In many cases improving one property comes at the expense of another. 

\begin{enumerate}[leftmargin=1.5em]
    \litem {High minimum row  distance $\rho$
    (between codewords)}
    With hard decoding (\ie nearest neighbor), the maximal number of prediction errors the scheme can recover from is 
    $\left\lfloor \left(\rho-1\right)/2 \right\rfloor$. 
    Using soft decoding, a high minimum distance is still vital for error correction, as seen from the error bound \eqref{eq:bound}.
    \litem {Low column correlation (between subproblems)}
    Intuitively, if two binary predictors often make errors on the same inputs,
    their mistakes become twice as hard to correct.
    Thus, uncorrelated columns (that yield uncorrelated binary subproblems) are generally considered advantageous.

    \litem {Efficient decoding algorithm}
    Traditionally ignored in many ECC works,
    the complexity of decoding prediction scores into codewords becomes essential in extreme classification tasks with thousands of  codewords or more.
    Recently, \citet{jasinska2016ltls} and
    \citet{evron2018wltls} utilized codebooks with a special structure to allow soft decoding using any decoding loss in a time complexity that depends only on the codebook width $\ell$ (which can be logarithmic in the number of codewords $K$).
    In contrast, exact soft decoding of arbitrary codebooks (\eg random or optimized ones)
    requires a time complexity at least linear in $K$. 
    % %

    \litem {Easy binary subproblems (low average loss $\varepsilon$)}
    The binary subproblems yield binary predictors with an average binary loss $\varepsilon$.
    The lower this loss is, the better the multiclass accuracy of the scheme becomes (see \eqref{eq:bound}).
    One way to lower $\varepsilon$ is to use high-capacity base learners (\eg kernel SVMs), but such rich models are often prone to overfitting or require more computation.

    A proper codebook design can lower  $\varepsilon$, 
    by making the subproblems \emph{easier},
    even for low-capacity learners.
    Following are design choices that can achieve this.
    \begin{enumerate}[leftmargin=1.7em]
        \litem{Sparse or imbalanced codebooks}
        \citet{allwein2000loss} extended the ECC scheme to ternary codes
        where $\M\!\in\!\left\{-1,0,+1\right\}^{K\times\ell}$.
        They showed that sparse columns generalize the one-vs-one scheme
        and that imbalanced columns generalize the \linebreak
        one-vs-all scheme.
        Both options can be seen as ways to create easier subproblems at the expense of the row distance or column correlation. 
        
        See \citet{zhou2016confusion}
        and Section~6 in \citet{allwein2000loss} for further discussion.

        \litem{Problem-dependent aspects} 
        Many papers design codebooks that are specifically suitable for the problem at hand while implicitly tuning the difficulty of the binary subproblems.
        
        Most of these works are guided by notions of class similarity.
        Some try (implicitly or explicitly) to create codebooks where similar classes have similar codewords (\eg \citealp{cisse2012compact}) in order to create easier subproblems.
        Others try the opposite {(\eg \citealp{martin2017factorization})} in order to enhance error correction between classes that are hard to separate, at the expense of harder subproblems.
        
        Notably, most methods
        balance preserving the similarity against
        other codebook properties
        (\eg the codeword distance between two very similar classes is encouraged to be $1$, whereas~$\rho$~is encouraged to be maximal).
        They create codebooks from scratch or alter existing ones.
        On the other hand, our observations next allow making \emph{predefined} codebooks
        more problem-dependent, 
        by simply assigning codewords to classes in an informed manner, and without harming other codebook properties which may be important.
    \end{enumerate}    
\end{enumerate}

\Section{Codeword-to-class assignments}
\label{sec:assignments_importance}
    
The error-correcting scheme 
implicitly assigns codewords to classes.
Both during training and during decoding, we arbitrarily assumed that the $k$th row in the codebook belongs to the $k$th class 
(see \eqref{eq:train_predictor} and \eqref{eq:soft-decoding}).
In an attempt to show robustness to codeword-to-class assignments,
\citet{dietterich1994solving} (Section~3.3.2 therein)
experimented on several random assignments and reported no significant accuracy variation.
However, they did not rule out the possibility that some assignments \emph{are} better than others.

We hypothesize that some assignments are \emph{significantly} better than others.
We first notice that given a codebook, 
different assignments induce different binary subproblems, potentially changing their difficulty 
and consequently the average binary loss $\varepsilon$.
Next, we define a scoring function that measures the extent to which close codewords are assigned to close classes.
This score later helps us conclude that \emph{similarity-preserving} assignments (\ie similar codewords to similar classes) are preferable.

\paragraph{Class-codeword score.}
Consider a class metric in the form of a distance matrix $\Dcls\!\!\in\!\!\doubleR^{K\times K}_{\ge 0}$.
For instance, 
$\Dcls$ can be (inversely proportional to) a symmetrized confusion matrix,
a matrix of distances between class embeddings,
or a matrix of distances between classes on a hierarchy tree.
Define the codeword distance matrix ${\Dcw\in\doubleR^{K\times K}_{\ge 0}}$ where
${\left(\Dcw\right)_{a,b}
    \triangleq
    \frac{1}{2}
    \norm{\M_{a,:}-\M_{b,:}}_1}$.
To account for the different scales of these matrices, we normalize them such that 
${\norm{\Dcls}_{\frob} = \norm{\Dcw}_{\frob} = 1}$.

Notice that an assignment corresponds to reordering, or permuting, the rows of the codebook $\M$ using a $K\!\times\!K$ permutation matrix $\Prm$.
Consequently, such an assignment corresponds to permuting the rows \emph{and} columns of the distance matrix $\Dcw$.

Given a codebook $\M$ and a class metric $\Dcls$.
We assess an assignment, or a permutation $\Prm$
of the rows in $\M$,
by defining the \emph{class-codeword score} as the Frobenius distance between $\Dcls$ and the permuted $\Dcw$:
\begin{align}
\label{eq:cc_dist}
    s_{\text{cc}}
    \left(\Prm\right)
    \triangleq
    \norm{\Dcls-
    \D_{\Prm\M}}_{\frob}
    =
    \norm{\Dcls-
    \Prm\Dcw\Prm^\top}_{\frob}
    .
    %
    %=
    %
    % \sum_{a,b\in\range{K}}
    % \left(
    %     \left(\Dcls\right)_{a,b}
    %     -
    %     \left(\Dcw\right)_{a,b}
    % \right)^2
\end{align}
Intuitively, an extreme case where $s_{\text{cc}}\left(\Prm\right)=0$
means that $\Dcls$ and the permuted $\Dcw$ completely ``agree", \ie similar codewords are assigned to similar classes, and dissimilar codewords are assigned to dissimilar classes
(realistically, given $\Dcls$ and $\Dcw$,
the minimum is often larger than zero).

\paragraph{Synthetic dataset.}

\appref{app:synthetic} illustrates some of the above ideas using a synthetic dataset.
For a specific codebook, we show that only \emph{one} assignment can perfectly fit the data, while \emph{all} other $(K!\!-1)$ assignments fail. %
\linebreak
Moreover, the only successful assignment assigns similar codewords to similar classes.
%
%Finally, one-vs-all (OVA) fails to fit the data.

\input{experiments.tex}

\input{related}

\Section{Conclusion}

Codeword-to-class assignments matter
because they vary greatly in the difficulty of subproblems they induce,
even for a \emph{predefined} codebook.
In classification tasks (of both small and large scales), similarity-preserving assignments lead to easier subproblems 
and better generalization performance.
Predefined codebooks can be advantageous when certain properties are crucial,
\eg specific minimum distance $\rho$ and number of predictors $\ell$, 
a given sparsity level,
or an efficient decoding algorithm.
Choosing an informed assignment 
according to class semantics,
allows for improving predefined codebooks by making them more problem-dependent.

Further research might discover that different usages require 
different assignment policies.
For instance, perhaps similarity-preserving assignments benefit generalization, 
while similarity-breaking assignments benefit robustness (see the discussion in \secref{sec:related}).

\subsubsection*{Acknowledgements}
We thank Koby Crammer and Thomas G. Dietterich for the fruitful discussions.
The research of DS was Funded by the European Union (ERC, A-B-C-Deep, 101039436). Views and opinions expressed are however those of the author only and do not necessarily reflect those of the European Union or the European Research Council Executive Agency (ERCEA). Neither the European Union nor the granting authority can be held responsible for them. DS also acknowledges the support of Schmidt Career Advancement Chair in AI.
\linebreak
Finally, we thank the Control Robotics \& Machine Learning (CRML) Lab at the Technion for their support.

\bibliographystyle{plainnat}
\bibliography{refs}

\clearpage

\onecolumn

\appendix

\aistatstitle{The Role of Codeword-to-Class Assignments in Error-Correcting Codes:
 \protect\\
 Supplementary Materials}

\input{supp.tex}

\end{document}

%% file: preamble.tex
\usepackage{xspace}
\usepackage{amssymb}
\usepackage{amsmath}
\usepackage{graphicx}
\usepackage{caption}
\usepackage{subcaption}
\usepackage{enumitem}
\newcommand\litem[1]{\item{\bfseries#1.\space}}

\usepackage{enumitem}

\usepackage[ruled]{algorithm2e}

\makeatletter
\renewcommand*{\algorithmcfname}{Algorithm Sketch}
\newcommand{\RemoveAlgoNumber}{\renewcommand{\fnum@algocf}{\AlCapSty{\AlCapFnt\algorithmcfname}}
\newcommand{\RevertAlgoNumber}{\algocf@resetfnum}}
\makeatother

\usepackage{booktabs}
\usepackage{subcaption}
\usepackage{url}
\usepackage{nicefrac}

\newlist{inlinelist}{enumerate*}{1}
\setlist*[inlinelist,1]{%
  label=(\roman*),
}

\usepackage{xcolor,soul}
\definecolor{itay}{RGB}{150,50,200}
\definecolor{todo}{RGB}{50,50,200}

\definecolor{red2}{RGB}{220,5,5}

\newcommand{\Norm}[1]{\left\lVert#1\right\rVert}
\newcommand{\norm}[1]{\Norm{#1}}
\newcommand{\bigO}[1]{\mathcal{O}\left({#1}\right)}

\DeclareMathOperator*{\argmin}{arg\,min}
\DeclareMathOperator*{\sgn}{sign}

\newcommand{\vect}[1]{\mathbf{#1}}

\newcommand{\mat}[1]{\mathbf{#1}}

\newcommand{\M}{\mat{M}}
\newcommand{\vx}{\vect{x}}
\newcommand{\vf}{\vect{f}}

\newcommand{\Prm}{\mat{P}}
\newcommand{\D}{\mat{D}}
\newcommand{\A}{\mat{A}}
\newcommand{\C}{\mat{C}}
\newcommand{\Dcls}{\D_{\text{cls}}}
\newcommand{\Ccls}{\C_{\text{cls}}}
\newcommand{\Dcw}{\D_{\M}}
\newcommand{\range}[1]{\left[{#1}\right]}
\newcommand{\doubleR}{\mathbb{R}}
\newcommand{\frob}{\text{F}}

\newcommand{\yeast}[0]{\texttt{yeast}\xspace}
\newcommand{\mnist}[0]{\texttt{MNIST}\xspace}
\newcommand{\cifarA}[0]{\texttt{CIFAR-10}\xspace}
\newcommand{\cifarB}[0]{\texttt{CIFAR-100}\xspace}

\newcommand{\lshtc}[0]{\texttt{LSHTC1}\xspace}
\newcommand{\dmoz}[0]{\texttt{LSHTC2}\xspace}
\newcommand{\ODP}[0]{\texttt{ODP}\xspace}
\newcommand{\odp}[0]{\texttt{ODP}\xspace}
\newcommand{\aloi}[0]{\texttt{aloi.bin}\xspace}
\newcommand{\wltls}[0]{\texttt{WLTLS}\xspace}
\newcommand{\arow}[0]{\texttt{AROW}\xspace}

\newcommand{\tsum}{{\sum}}

\newcommand{\ie}{i.e.,\xspace}
\newcommand{\eg}{e.g.,\xspace}

\def\secref#1{Section~\ref{#1}}
\def\appref#1{App.~\ref{#1}}
\def\figref#1{Figure~\ref{#1}}
\def\tabref#1{Table~\ref{#1}}

\def\Section#1{\section{\uppercase{#1}}}

\newcommand\smalldots{.\hskip1.7pt\!.\hskip1.7pt\!.\hskip1pt}

%% file: experiments.tex
\Section{Experiments}
\label{sec:experiments}

We test our hypothesis and demonstrate the validity of our claims in two regimes.
First, in \secref{sec:exp-exhaustive} we run extensive experiments on small datasets and illustrate how codeword-to-class assignments vary greatly in their accuracy. 
We show that this variation is mostly explained by the average binary loss $\varepsilon$ 
from \eqref{eq:average_binary_loss}
and the class-codeword score
from \eqref{eq:cc_dist}.
We conclude that 
similarity-preserving 
assignments are vital for inducing easy binary subproblems.
Then, in \secref{sec:extreme_experiments} we 
employ similarity-preserving assignments on codebooks for extreme classification.
\linebreak
We show how the structure of specific predefined codebooks facilitates finding good assignments and improve performance on datasets with up to 104K classes.

\subsection{Exhaustive Experiments}
\label{sec:exp-exhaustive}
\paragraph{Datasets.}
We start by testing our hypothesis on $3$ small datasets with $K=10$ classes: \mnist \citep{lecun1998mnist},
\cifarA \citep{krizhevsky2009cifar},
and \yeast \citep{Dua2019UCI}.

\begin{table}[h!]
  \centering
  \caption{
  Exhaustive Experiments' Datasets
  \label{tbl:datasets}
  }
  %\vskip 1mm
  \begin{tabular}{l|l|lll|l}
    \toprule 
    Dataset & Area 
    & Feat.
    & Train
    & Test
    & Model
    \\
    \midrule
    %\\
    \mnist & Vision & 
    784
    &
    60K
    &
    10K
    &
    Linear
    \\
    \cifarA & Vision &
    3,072
    & 
    50K
    &
    10K
    &
    Linear
    \\
    \yeast & Life & 
    8 & 
    1,284
    &
    200
    &
    DT
    \\
    \bottomrule
  \end{tabular}
\end{table}

\paragraph{Codebooks.}
We experiment on 3 predefined codebooks:
Two random dense codebooks (generated like in \citealp{allwein2000loss}) of widths $\ell=8,15$ having row distances of $\rho=3,5$ (respectively)
and a truncated Hadamard matrix (see \citealp{hoffer2018fix}) with $\ell=15$ and $\rho=8$.

\paragraph{Experimental setup.}
Working with only $K=10$ classes allows us to extensively validate our claims on \textbf{all} possible $K!\approx 3.6M$ assignments
of each combination of a dataset and a predefined codebook.
Notice that given such a combination, we need not \emph{train} $K!$ assignments from scratch.
Instead, we train only $2^{K-1}\!=\!512$ binary predictors and construct every possible assignment from them.
\linebreak
This technique saves time and decreases the variance of the evaluated test accuracy 
(details in \appref{app:creating_assignments}).

To demonstrate the flexibility of our observations,
we use two different base learners.
For \mnist and \cifarA,
we train $\ell$ linear predictors using the (soft-margin) SVM algorithm.
For \yeast, each binary predictor is a decision tree 
(built by the Gini splitting criterion and a minimum of 3 samples to split a node).
Hyperparameters were tuned using cross-validation (details in \appref{app:tuning}).

In the decoding step \eqref{eq:soft-decoding},
we use the hinge loss, 
corresponding also to the loss minimized by the SVM
used for training the linear base learners.

\subsubsection{Variation in performance of assignments}
\label{sec:variability}

\begin{figure}[h!]
    \centering
    \includegraphics[width=1\linewidth]{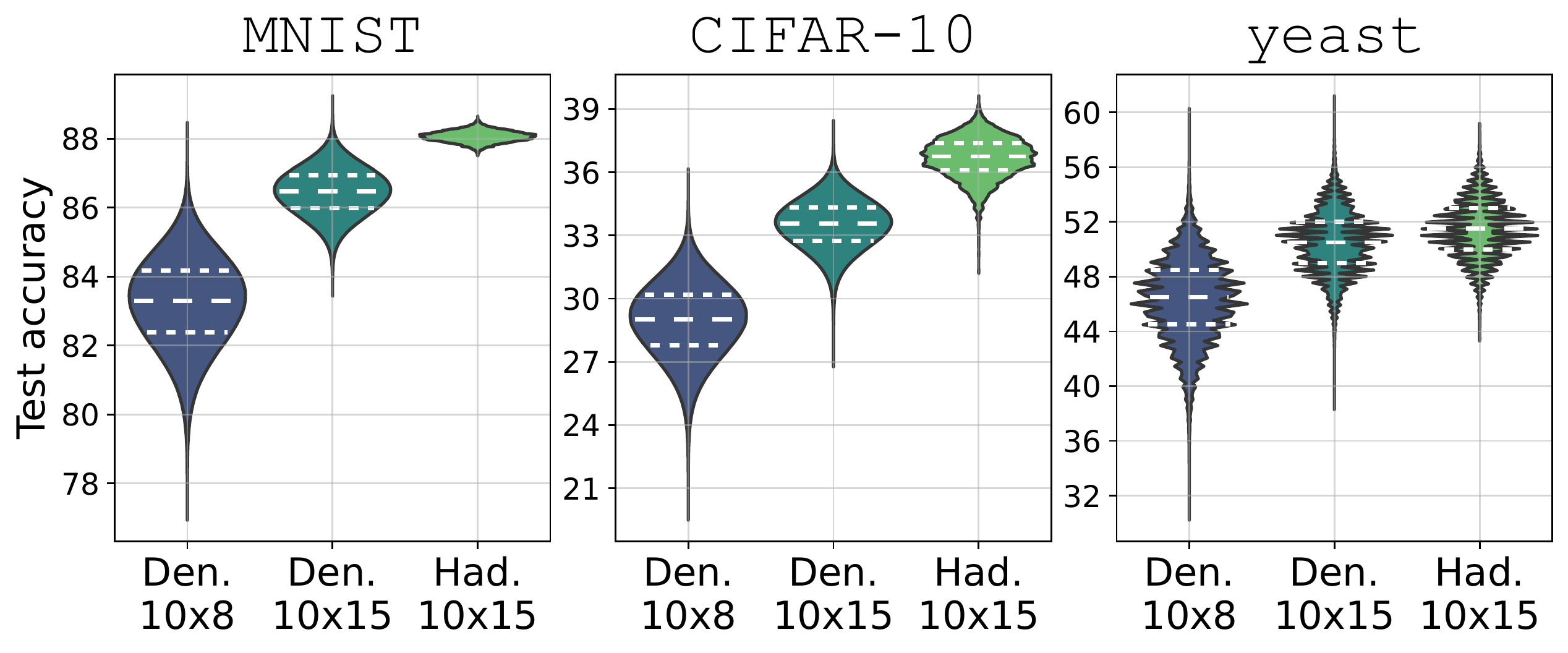}
    \vskip -1.5mm
    \caption{
    %\small
    Variation of test accuracy across all
    $10!\approx 3.6M$ assignments of 3 codebooks on 3 datasets.
    Dashed lines indicate quartiles (except where the plot is too narrow).
    \linebreak
    There is a large variation in performance across different assignments of the same codebooks.
    }
    \label{fig:variation}
\end{figure}

\figref{fig:variation} illustrates the large variation in performance for different assignments of given codebooks.
For instance, %\figref{fig:variation_mnist}
in \mnist we observe that using the random dense codebook of width $\ell=8$, 
the worst assignment achieves $\approx77\%$ test accuracy,
while the best assignment achieves $\approx88.5\%$.

\begin{figure*}[t!]
    \centering
    \begin{subfigure}{.99\textwidth}
      \centering
      \includegraphics[width=.99\linewidth]{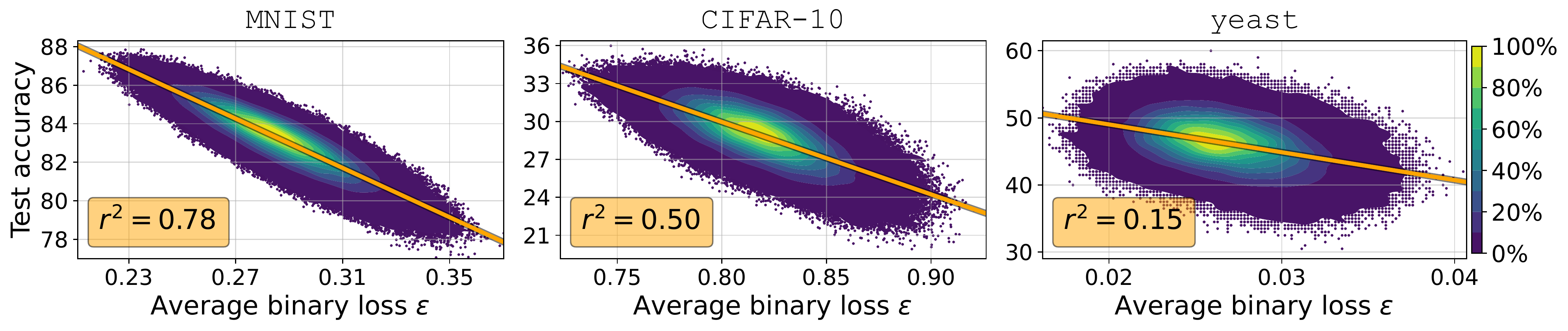}
    %   \caption{\mnist: $r^2=0.76$%
    %   \label{fig:error_mnist}}
    \vskip -1mm
        \caption{Correlation between average binary (train) loss and test accuracy.
        Assignments that induce easier subproblems perform better.}
    \label{fig:average_loss}
    \end{subfigure}
    \vskip 1mm
    \begin{subfigure}{.99\textwidth}
      \centering
      \includegraphics[width=.99\linewidth]{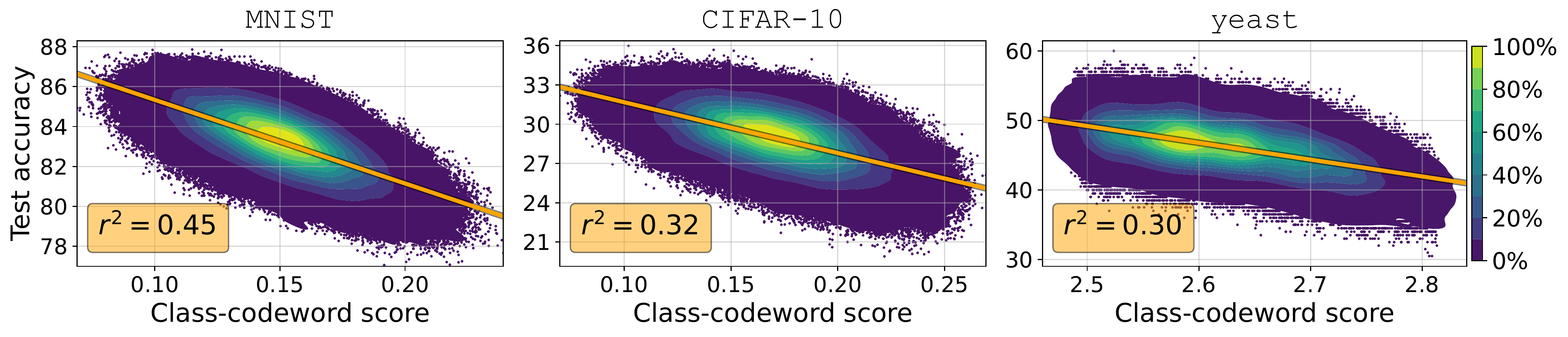}
    \vskip -1mm
    \caption{
    Correlation between class-codeword score (created by confusion matrices).
    Similarity-preserving assignments
    perform better.
    }
    \label{fig:class_codeword}
    \end{subfigure}
    \vskip -1mm
    \caption{
    \label{fig:empirical_combined}
    \small
    The empirical distributions of \emph{all} the 3.6M assignments of the random $10 \!\times\! 8$ codebook
    on the 3 datasets.
    \linebreak
    \textbf{Top:}
    Test accuracy vs. average binary (train) loss from \eqref{eq:average_binary_loss}.
    \textbf{Bottom:} 
    Test accuracy vs. the class-codeword score from \eqref{eq:cc_dist}.
    \linebreak
    Each level set contains $\approx\! 10\%$ of the assignments.
    The $10^{-3}$ least probable assignments are scattered
    as individual points.
    \linebreak
    Regressors computed on all assignments are plotted in orange.
    Also written are the coefficients of determination ($r^2$).
    }
    %\vskip -2mm
\end{figure*}

In all 3 datasets, the narrow ($\ell < K$) codebook exhibits higher variation in performance.
This can be explained by the low minimum distance ($\rho=3$) which does not allow for meaningful error correction,
making the average binary loss $\varepsilon$ a more dominant factor in performance.

\paragraph{Equidistant codebooks.}
The low variation in the Hadamard codebook (especially in
\mnist)
probably stems from it being an \emph{equidistant codebook} (every two codewords are in the same distance from each other).
In such codebooks, the class-codeword score \eqref{eq:cc_dist} 
remains constant across all assignments
(since $\forall\Prm\!: \Dcw = \Prm\Dcw\Prm^\top$).
This also supports the following findings
(\secref{sec:exp_class_similarity})
that the class-codeword score is a lead factor in the observed performance variation.

\subsubsection{
Some assignments induce easier subproblems}
\label{sec:exp_train_loss}

\figref{fig:average_loss}
shows the correlation between the average binary train loss $\varepsilon$ and the test accuracy.
We plot the empirical distribution (using kernel density estimation) of all 3.6M assignments ran on the 3 datasets using the $10\!\times\!8$ random dense codebook.

For \mnist (top~left), the correlation between the test accuracy and $\varepsilon$ is the highest ($r^2 = 0.78$).
The other two datasets exhibit lower correlations, 
but large performance gaps are still explained by $\varepsilon$ which roughly quantifies the difficulty of subproblems induced by each assignment.

We observe a similar behavior in another $10\!\times\! 8$ codebook
and a wider $10\!\times\! 15$ codebook
as well (\appref{app:correlation}).

The observed correlation between performance and the average binary loss $\varepsilon$ is itself not surprising and can be expected from the error bound in \eqref{eq:bound}.
However, our results stress that different \emph{assignments} of the \emph{same} codebook induce binary subproblems of different difficulty.

\pagebreak

\subsubsection{Similarity-preserving assignments are better}
\label{sec:exp_class_similarity}

We now test the effect of class similarity on an assignment's performance.
We use the class-codeword score \eqref{eq:cc_dist}
to assess how close are codewords of similar classes.

\paragraph{Sources of class similarity.}
Our class-codeword score %\eqref{eq:cc_dist}
requires a matrix $\Dcls$ corresponding to a class metric.
Here, we use two \emph{different} class metrics to strengthen our findings.
%
%\linebreak
%
First, we use the (training) confusion matrices of one-vs-all predictors,
assuming that confusable classes are semantically similar 
(a common assumption; see \citealp{zhou2016confusion}).
Then, in \appref{app:correlation}, we use Euclidean distances between the means of raw features of each class.
\linebreak
\appref{app:class_metrics} explains how we turn a confusion matrix (a similarity matrix) into a distance matrix.

\paragraph{Results.}
\figref{fig:class_codeword} shows the correlation between our class-codeword score %\eqref{eq:cc_dist} 
and test accuracy.
We use the same random dense $10\!\times\!8$ codebook as before,
and compute the class-codeword score 
from confusion matrices
(see above).

For example, 
the plot on the bottom-middle
shows the distribution of all 3.6M assignments ran on \cifarA. 
On average,
assigning similar codewords to similar classes (thus minimizing the class-codeword score) 
 improves the test accuracy from
$\approx\! 29\%$ to 
$\approx\! 32.5\%$. % 
Moreover, assigning similar codewords to \emph{dissimilar} classes evidently worsens the performance significantly
(to $\approx\! 25.5\%$)

We observe a similar behavior in another $10\!\times\! 8$ codebook
and a wider $10\!\times\! 15$ codebook
as well (\appref{app:correlation}).

\pagebreak

\subsubsection{Summary}
Some assignments of 
\emph{the same} codebook induce much easier binary subproblems than others do.
Our class-codeword score largely explains the performance of an assignment.

Computing the class-codeword score of one assignment is cheap and mainly requires calculating the distance between two $K \times K$ matrices.
Thus, when $K=10$, exhaustively iterating 
\emph{all} 3.6M assignments
to find the one minimizing that score,
takes only a few minutes on a single CPU. 
Overall, a similarity-preserving assignment found exhaustively \emph{before} training should yield a much better test accuracy than a random assignment.

In \appref{app:cifar100} we show that 
the class-codeword score also controls performance in a larger dataset
(\cifarB), 
where any exhaustive experiment becomes intractable.
We demonstrate that
similarity-preserving assignments,
originating from the distances between {\tt fastText}
embeddings of \emph{class names},
significantly improve performance.

\subsection{Extreme Multiclass Classification (XMC)}
\label{sec:extreme_experiments}

\begin{figure*}[ht]
    \centering
    \begin{subfigure}{.255\textwidth}
      \centering
      \caption{\aloi ($K\!\!=\!1,\!000$)}
      \label{fig:aloi}
      \vskip -2mm
      \includegraphics[width=.99\linewidth]
      {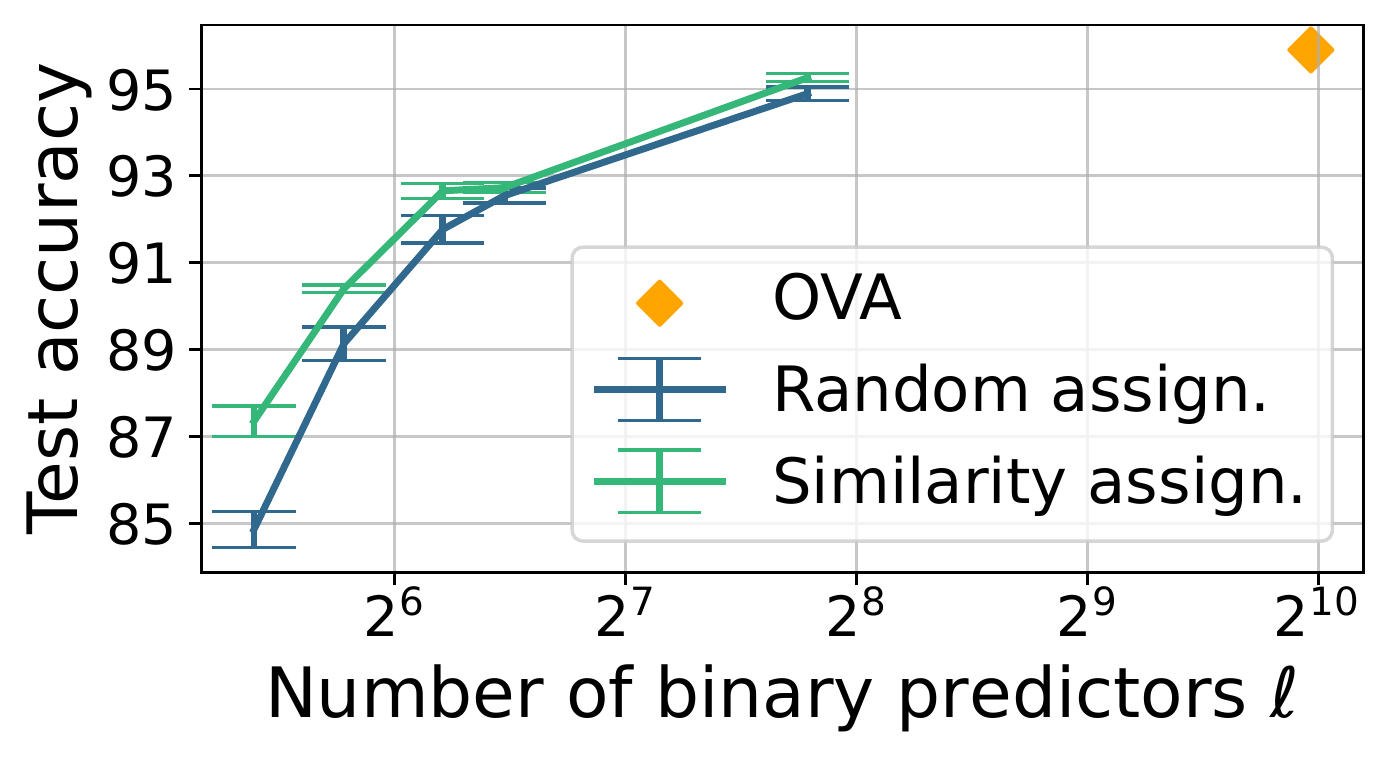}
    \end{subfigure}
    %
    %\hspace{.005\textwidth}
    \begin{subfigure}{.24\textwidth}
      \centering
      \caption{\lshtc ($K\!\!=\!12,\!294$)}
      \label{fig:lshtc1}
      \vskip -2mm
      \includegraphics[width=.99\linewidth]
      {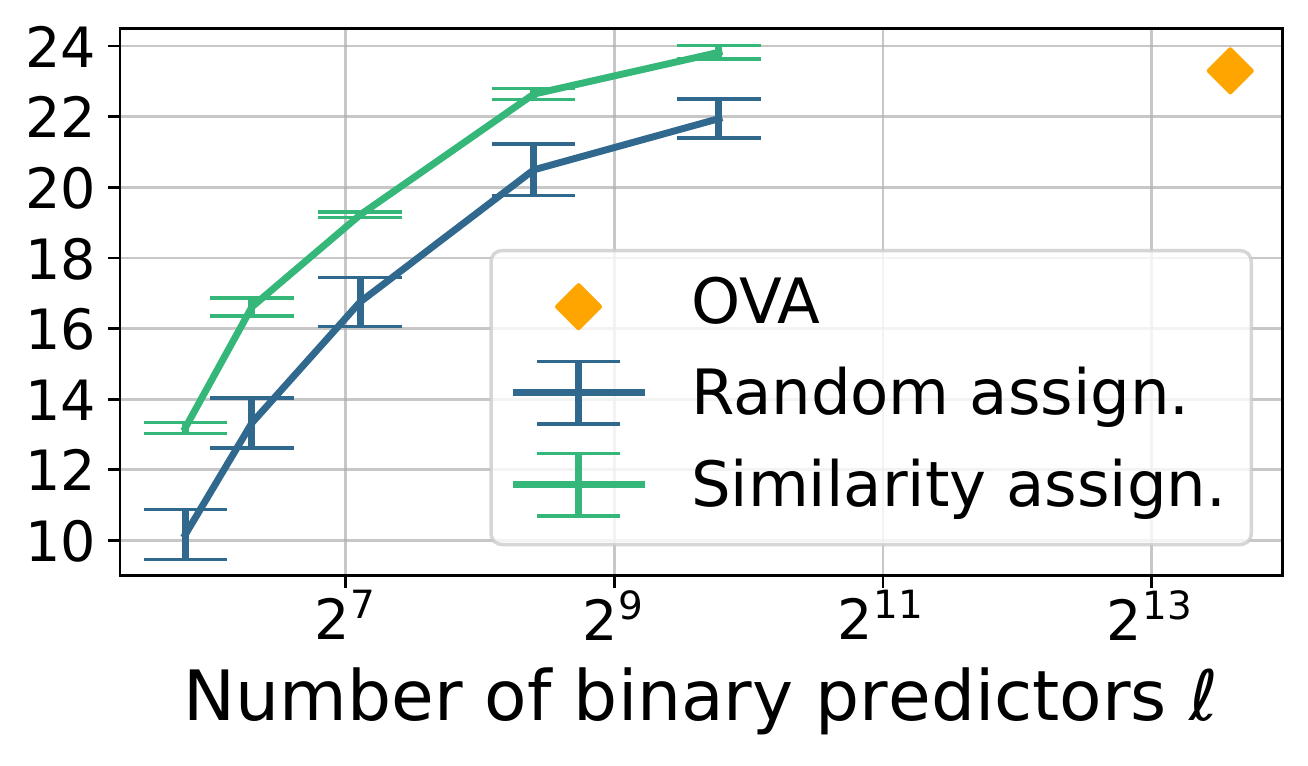}
    \end{subfigure}
    %
    %\hspace{.005\textwidth}
    \begin{subfigure}{.24\textwidth}
      \centering
      \caption{\dmoz ($K\!\!=\!27,\!840$)}
      \label{fig:lshtc2}
      \vskip -2mm
      \includegraphics[width=.99\linewidth]
      {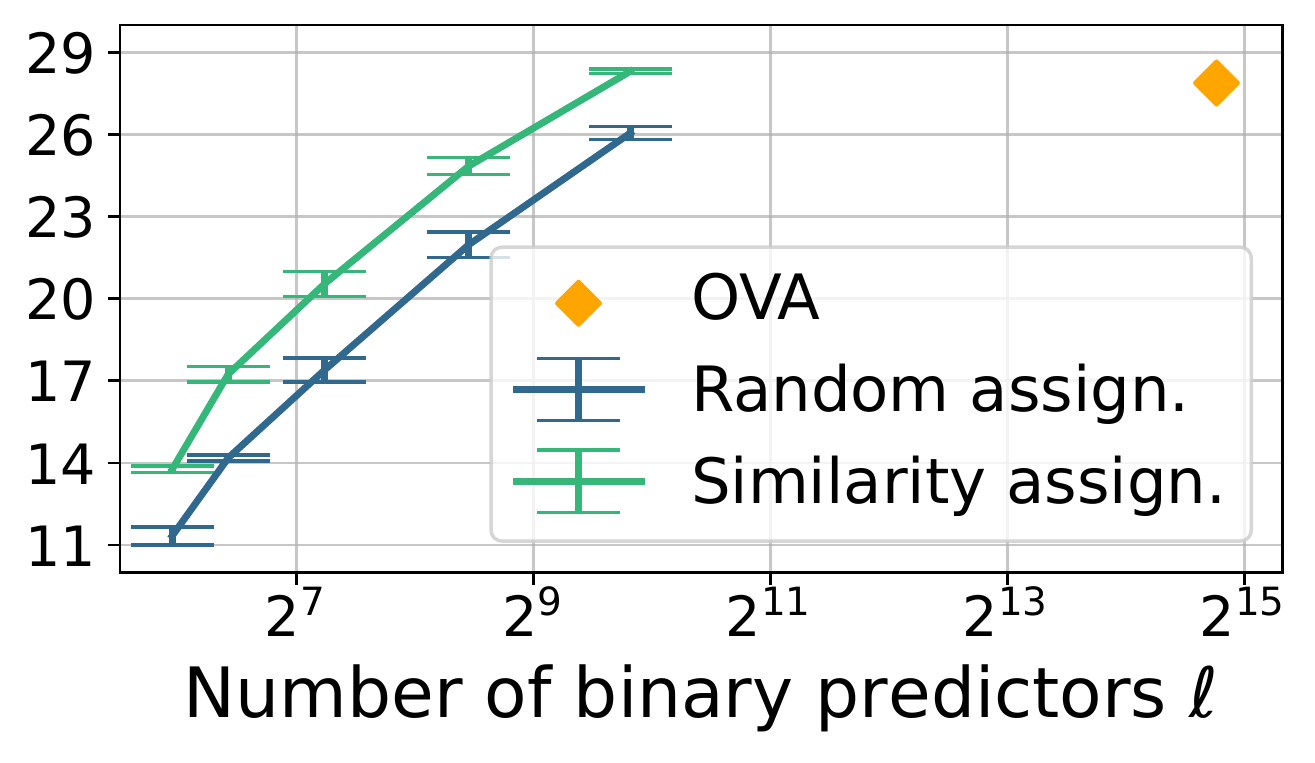}
    \end{subfigure}
    %\hspace{.005\textwidth}
    \begin{subfigure}{.24\textwidth}
      \centering
      \caption{\ODP ($K\!\!=\!104,\!136$)}
      \label{fig:odp}
      \vskip -2mm
      \includegraphics[width=.99\linewidth]
      {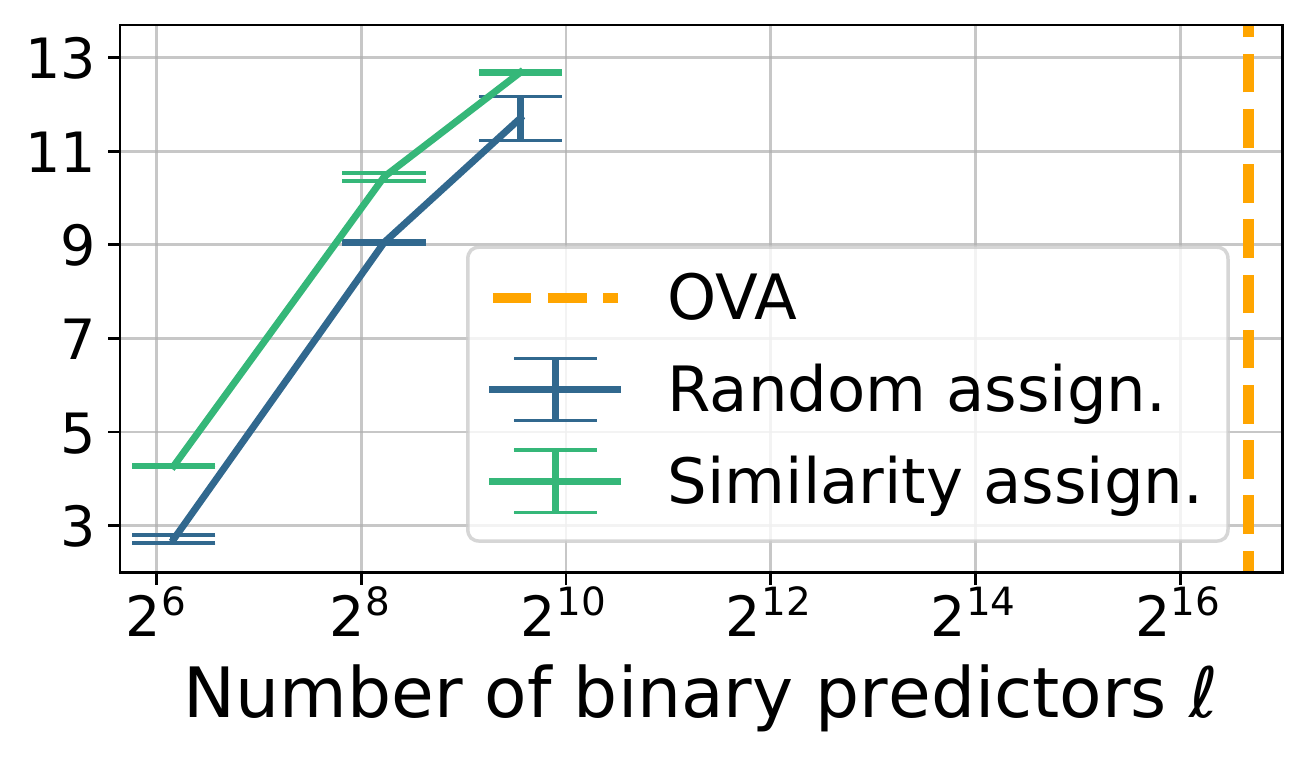}
    \end{subfigure}
    \vskip -2mm
    \caption{
    \small
    Results on extreme datasets.
    We run \wltls codebooks using different predictor numbers $\ell$.
    Errorbars indicate $\pm$2 empirical standard deviations of $5$ runs).
    Results are available in a tabular form in \appref{app:tabular}.
    Due to computational infeasibility, 
    we do not report the performance of one-vs-all (OVA) on the largest dataset (\dmoz), but just mark its number of binary predictors instead (where $\ell\!=\!K$).
    In all datasets, assigning similar codewords to similar classes significantly improves performance.
    }
    \label{fig:lshtc}
    \vskip -2mm
\end{figure*}

We now utilize our understanding that similar codewords should be assigned to similar classes on four XMC benchmarks trained using XMC-dedicated codebooks.
We show that in the extreme regime as well --- similarity-preserving assignments are significantly better than random ones.

\paragraph{Datasets.} 
We experiment on four XMC preprocessed benchmarks --
\lshtc, \dmoz \citep{Partalas2015LSHTC},
\aloi \citep{rocha2013multiclass,yen2016pdsparse},
and \ODP \citep{bennett2009refined}.
The datasets are described briefly below and in detail in  
\appref{app:xmc}.

\begin{table}[h!]
  \centering
  \caption{
  %\small
  Extreme Benchmarks.
  Further details in \appref{app:xmc}.
  \label{tbl:extreme_datasets}
  }
  %\vskip 1mm
  \vspace{-1mm}
  \begin{tabular}{l|l|ll|l}
    \toprule 
    Dataset & Area & Classes & Features
    & Similarity
    \\
    \midrule
    %\\
    \tt{aloi} & Vision & 
    1K &
    637K
    &
    Clustering
    \\
    \lshtc & Text &
    12K
    & 1.2M
    &
    Given
    \\
    \dmoz & Text & 
    27K & 
    575K 
    &
    Given
    \\
    \ODP & Text & 
    104K & 
    423K &
    Clustering
    \\
    \bottomrule
  \end{tabular}
\end{table}

\vspace{-2mm}

\paragraph{Sources of class similarity.}
For all datasets, our algorithm below uses class taxonomies given in a form of a tree. 
These taxonomies are either known in advance (in \lshtc and \dmoz) 
\emph{or computed} by a simple hierarchical clustering algorithm on class means (in \aloi and \odp).
Again, using multiple sources of class similarities corroborates the soundness of our findings below.

\paragraph{Experimental setup.}
We use the code from the publicly available repository of \citet{evron2018wltls} to learn using their \wltls codebooks.
To use our similarity-preserving codeword-to-class assignments,
we edit their scripts to allow for fixed assignments (rather than random ones).%
\footnote{
The updated GitHub repository is available on
\linebreak
\url{https://github.com/ievron/wltls}
}
\linebreak
We also use the same learning setup --- 
as a base learner, we use \arow \citep{crammer2009arow}, which is an online algorithm for learning linear classifiers,
and we also use the exponential loss for the soft decoding step in \eqref{eq:soft-decoding}.
We run all experiments sequentially on a single \texttt{i7} CPU. 
In practice, each binary predictor can be trained on a separate CPU.

For each dataset, we train several \wltls codebooks of various widths $\ell$.
Each codebook is learned 5 times using random assignments and 5 times using similarity-preserving assignments,
found as described below
(here, randomness stems from shuffling the training set).

For comparison, we also train one-vs-all (OVA) models using the same base learner -- \arow.
Our OVA results are better than the ones reported in \citet{evron2018wltls}, since we apply oversampling \citep{ling1998oversampling} to overcome the high imbalance in each OVA subproblem.

\paragraph{Finding similarity-preserving assignments.}
We exploit the graph structure of \wltls codebooks which embed $K$ codewords on source-to-target paths of a directed acyclic graph (DAG) with exactly $K$ such paths.
Since the class taxonomies are also DAGs, a quick-and-simple algorithm arises for assigning similar codewords to similar classes.

\RemoveAlgoNumber

\vskip -1mm

\begin{algorithm}
    \footnotesize
    \begin{minipage}{.92\linewidth}
    \vskip 1mm
    \textbf{Input:}
    \BlankLine
    \begin{enumerate}[leftmargin=5mm,topsep=2pt,]\itemsep.1pt
        \itemsep0em 
        \item The dataset's class taxonomy (given or learned)
        \item A \wltls coding DAG suitable for $K$ classes
    \end{enumerate}
    \vskip 1mm
    \textbf{Algorithm:}
    \BlankLine
    \begin{enumerate}[leftmargin=5mm,topsep=2pt,]\itemsep.1pt
        \item Traverse the class tree with DFS to obtain an ordering $\left(a_1,\smalldots,a_K\right)$ of leaves (\ie classes);
        \item 
        Recursively iterate \emph{all} $K$ paths in the coding DAG, 
        to obtain an ordering $\left(b_1,\smalldots,b_K\right)$ of paths (\ie codewords);
        \item Assign class $a_i$ to codeword $b_i$.
    \end{enumerate}
    \end{minipage}
    \caption{\footnotesize
    Naive assignment for \wltls}
    \vskip 1mm
\end{algorithm}

\vskip -1mm

The proposed algorithm preserves similarities by assigning similar classes to similar codewords.
Intuitively, in most cases classes $a_i$ and $a_{i+1}$ are close on the taxonomy 
and paths $b_i$ and $b_{i+1}$ are similar on the codebook's DAG.
\linebreak
We illustrate this algorithm in
\appref{app:extreme_alg_details}.

Despite its simplicity, the algorithm finds assignments with exceptionally low class-codeword scores \eqref{eq:cc_dist} compared to the scores of random assignments.
For example, for the smallest codebook of \lshtc ($\ell\!=\! 56$), random assignments 
exhibit an average score of 
$\widehat{s_{\text{cc}}}\!\approx\! 0.061$ with an empirical standard deviation of $1.33\cdot10^{-5}$;
while the assignment our algorithm finds has a score of $s_{\text{cc}}\!\approx\!0.049$.
That is, compared to random assignments, 
our algorithm decreases the score by more than $900$ standard deviations~(!).

\pagebreak

\paragraph{Results.}

\figref{fig:lshtc} demonstrates the advantage of similarity-preserving codeword-to-class assignments.
For each dataset, we compare the test accuracy of random assignments to that of similarity-preserving assignments
across various codebook widths $\ell$.

We plot the test accuracy averages of the 5 runs of each combination of a codebook width and an assignment method,
accompanied by 2 empirical standard deviations
(full result tables are given in \appref{app:tabular}).
\linebreak
In almost all cases, similarity-preserving assignments 
lead to a statistically-significant improvement over random assignments.
Moreover, in 16 out of 18 cases, similarity-preserving assignments exhibit a lower variance.
\linebreak
In \lshtc and \dmoz, 
similarity-preserving assignments make the codebooks competitive with OVA while training up to 32 times fewer predictors.

In the two larger codebooks of \aloi,
our assignments do not improve much over random ones. This probably happens because when $\ell$ approaches $K$,
the underlying \wltls codebooks become almost equidistant.

\paragraph{Summary.}
Similarity-preserving assignments significantly improve 
codebooks dedicated to extreme classification.
By exploiting class semantics,
such assignments turn \emph{predefined} codebooks with regime-specific advantages (\eg fast decoding algorithms) into problem-dependent codebooks, without losing those advantages.

%% file: related.tex
\Section{Related work}
\label{sec:related}

Our work is of a retrospective nature and calls for an elaborate discussion of its connections with decades of existing research on error-correcting codes.

Codebooks with easy subproblems are obviously preferable.
\citet{bai2016ecoc} design a codebook by selecting a subset of the easiest columns
out of all possible columns.
They exhaustively train on \emph{all} these columns and select a column subset based on the trained predictors' accuracy.
This works well but does not scale gracefully
(\eg for merely $K\!=\!10$ classes, it requires \emph{training} $2^{K-1}\!=\!512$ predictors).
Instead, many works (including ours) exploit extra knowledge on classes
to create easy subproblems.

\paragraph{Codebook design methods.}
While we point out that similarity-preserving assignments improve a \emph{predefined} codebook by making it \emph{problem dependent}, 
most works try to design 
the \emph{entire} codebook.
Given a dataset, designing optimal codebooks is a hard problem due to their discrete nature \citep{crammer2002learnability}.
As a remedy, some papers take greedy approaches,
\eg sequentially adding optimized columns \citep{pujol2008incremental}
or solving integer programming formulations
\citep{gupta2020integer,gupta2022ECOCcoloring};
while others take approximate approaches,
like solving relaxed continuous optimization problems (\eg \citealp{zhang2009spectral, rodriguez2018beyond}).

\paragraph{The class-similarity controversy.}
Many papers incorporate different notions of class similarity into
their design process. 
Interestingly, some encode similar classes with \emph{similar} codewords
\citep{zhang2009spectral, 
cisse2012compact,zhao2013sparse,zhou2016confusion,
rodriguez2018beyond,mcvay2020generalization},
whereas others encode similar classes with \emph{dissimilar} codewords
\citep{pujol2008incremental,martin2017factorization,jaiswal2020mute,gupta2020integer,wan2022efficient}.
For instance,
\citet{martin2017factorization}
look for a codebook ${\M\!\in\!\left\{-1,+1\right\}^{K\times\ell}}$
that \emph{minimizes}
${\norm{\Dcls
-
\M\M^\top
}_{\frob}^2}$,
while balancing against other codebook properties. 
%(\secref{sec:codebook_properties}).
%
In fact, they \emph{maximize} our score \eqref{eq:cc_dist} instead of minimizing it,
since 
${\M\M^\top = \ell \vect{1}_{K\times K} 
-
\Dcw}$.

Existing literature on adversarial robustness has thus far considered assigning \emph{dissimilar} codewords to similar classes 
(\eg \citet{gupta2020integer,wan2022efficient}).
in order to improve the error-correcting capabilities between easily-confusable classes, especially in the presence of an adversary.
On the other hand, our study shows that similarity-preserving assignments improve the separability and classification performance in traditional settings.
\linebreak
An interesting future direction should be to perform adequate ablation studies in the adversarial learning regime and examine the tradeoff between separability (maximized by similarity-preserving assignments) and robustness (maximized by similarity-breaking ones).

\paragraph{Class similarity in  extreme classification (XMC).}
In~\secref{sec:extreme_experiments} 
we use a class taxonomy
to improve a codebook that requires training very few predictors compared to one-vs-all or hierarchical models.
A closely related work \citep{cisse2012compact} designs XMC-codebooks using a learned class-similarity. 
However, their codebooks do not allow fast decoding like the ones we use.
Other related approaches learn hierarchical models using a given (or learned) class taxonomy, to either benefit from a $\bigO{\log K}$ prediction time \citep{bengio2010labeltrees},
or to alleviate the computation of the softmax while training a deep network
\citep{morin2005hierarchical}.
Another approach directly builds a codebook from a class taxonomy \citep{pujol2006decoc}.
However, these approaches  train $\bigO{K}$ predictors, implying longer training and \emph{linear} space requirements.
Recently, \citet{mittal2021decaf} incorporated label metadata in the training of \emph{deep} extreme classification models (much larger than the linear \wltls models 
we use).
Finally, \citet{rahman2018zeroshot} use class semantics to improve zero-shot performance,
which may be relevant to XMC tasks which often suffer from a long tail of classes \citep{babbar2014power},
some having few to no training examples.

\paragraph{Ordinal classification and regression tasks} 
can also be tackled with ECC.
Interestingly, successful assignments used implicitly in these areas
often follow a similar rule-of-thumb like we do --
they encode target labels that are similar 
(\ie close on the real line)
using similar codewords.
%\linebreak
%
For instance, see the {\tt Unary} and {\tt HEXJ}
codebooks in \citet{shah2022regression} 
(the first codebook is equivalent to the underlying codebook in 
\citealp{li2006ordinal}) or the random ordered-splits in \citet{huhn2008ordinal}.
However, similarities in these areas 
(\ie distances on the real line) are much simpler than the inter-class relations examined in our paper.

\paragraph{Nested dichotomies (ND)} offer another reduction from multiclass tasks to binary ones. 
Basically, ND models split classes recursively in a binary hierarchical structure, where each tree node corresponds to a binary classification subproblem.
One could either use a single tree \citep{fox1997applied} or an ensemble of  trees 
\citep{frank2004ensembles}.
%
%In both cases, 
The resulting models can be seen as a special case of ECC.

\citet{melnikov2018effectiveness}
conduct an experiment that is closely related to our variation demonstration in \secref{sec:variability}.
They show that the assignment of classes to leaves of a \emph{single} ND tree greatly affects the model's performance, and report a high variation in the performance of \emph{randomly-sampled} NDs 
(the tree structure was also shown to be important in \citealp{mnih2008scalable}).
However, their tree corresponds to a codebook with a minimum Hamming distance of $\rho\!=\!1$ 
(\ie a prediction mistake in \emph{one} inner node necessarily results in a multiclass error).
Thus, it is not immediate that their findings generalize to codebooks with  higher error-correcting capabilities (like the ones we use).
\linebreak
Importantly,
we do not only point out 
the performance variation of codeword-to-class assignments,
but also clearly show it is explained by class-similarity (\secref{sec:exp_class_similarity}).

\paragraph{Model capacity.}
Codeword-to-class assignments control the difficulty of the binary subproblems, which is 
\linebreak 
naturally more crucial when the base learners are weaker 
(see the ${\varepsilon}/{\rho}$ factor in \eqref{eq:bound}).
Related phenomena have been exhibited in
ordinal classification \citep{huhn2008ordinal}
and nested dichotomies \citep{melnikov2018effectiveness}
as well.
In this paper, we demonstrated our findings using 
relatively weak
linear models and decision trees over raw features
(\secref{sec:exp-exhaustive})
and
preprocessed ones (\secref{sec:extreme_experiments}; \appref{app:cifar100}).

Even high-capacity models like neural networks 
are likely to favor
similarity-preserving assignments.
\citet{zhai2018classificationForMetric} show that 
a deep classification network
implicitly performs \emph{metric learning} ---
training embeds the classes' weight vectors in the last %fully connected 
linear layer (preceding the softmax) in a way that reflects underlying class semantics
(see also \citet{kusupati2021llc}).
Similarity-preserving codebooks can be seen as fixing the last layer using a matrix that already reflects such semantics at initialization (see also 
Sec.~3.3
in \citealp{hoffer2018fix}).

Notably, complex models can 
attain a very low average training binary loss $\varepsilon$ such that the training error bound~\eqref{eq:bound} becomes $< {1}/{m}$, implying no \emph{training} mistakes.
\linebreak
However, 
this does not make assignments unimportant.
\linebreak
If, for example, we train and decode using an exponential loss, then complex learners can obtain an extremely low loss $\varepsilon$, but never $0$.
In such cases, similarity-preserving assignments should \emph{still} yield a lower $\varepsilon$.
In turn, a lower training loss, even when the error is already $0$, is linked to better generalization,
both theoretically and practically 
(\eg \citet{soudry2018journal}).

\paragraph{Limitations of design methods.}
Similarity-preserving assignments can enhance almost \emph{any} predefined codebook,
while design methods are often restricted to codebooks with certain properties.
%
%\linebreak
%
For instance, the spectral method \citep{zhang2009spectral} creates only narrow codebooks
(where $\ell\!\le\! K$)
and does not explicitly take the minimum row distance $\rho$ into account, which may not be best suited for small datasets
(\eg on \cifarA with ${\ell\!=\!8}$, their method yielded two identical rows).
Other methods scale poorly with the number of classes \citep{bai2016ecoc, escalera2008subclass}.
Some are more suitable for creating balanced dense columns
\citep{zhang2009spectral,rodriguez2018beyond}
while others focus on sparse columns \citep{pujol2006decoc}.

\paragraph{Limitations of finding informed assignments.}
Designing problem-dependent codebooks from scratch is naturally more flexible than only assigning classes to predefined ones.
Objective scores can be optimized more freely when the codebook itself is not fixed like in predefined codebooks.
However, predefined codebooks can have favorable properties like fast decoding algorithms, hence it is important to be able to find informed assignments for them.

We use our class-codeword score
${
    %s_{\text{cc}}
    % \left(\Prm\right)
    % \triangleq
    \norm{\Dcls
    \!-
    \Prm\Dcw\Prm^\top}_{\frob}}$ 
%in~\eqref{eq:cc_dist}
mainly to demonstrate the
superiority of similarity-preserving assignments (\secref{sec:exp_class_similarity}).
One could also employ this score as a surrogate to control the difficulty of subproblems, and directly minimize it on a given codebook
to find an optimal similarity-preserving
assignment.
However, finding this optimum corresponds to solving a weighted graph-matching problem,
which does not have a known efficient  
algorithm
\citep{umeyama1988WGMP}.
Instead, one could settle for assignments with a low (but possibly sub-optimal) score.
We~exemplify this using a local search on a \cifarB codebook (\appref{app:cifar100}).
An exception where our score is constant and assignments are less impactful, is in equidistant codebooks (\eg Hadamard, OVA, OVO;
see \secref{sec:variability}).
This suggests that equidistant codebooks are perhaps more suitable when no class semantics are available.
They can also be expected to yield smaller variation
(see \figref{fig:variation}).
See also \citet{james1998errorPICT} who linked such codebooks to Bayes optimality.
As a downside, these codebooks must be wide 
($\ell\ge K$), which is unacceptable in many cases such as extreme classification.

\paragraph{Greedy assignment policies.}
After submitting our paper, we became aware of two recent works closely related to ours
that also improve the performance of a given codebook using codeword-to-class assignments.
\linebreak
\citet{mcvay2020generalization} exploits a sparse class-similarity matrix to greedily assign similar codewords to similar classes.
\linebreak
\citet{wan2022efficient} employ ECC for adversarial learning,
by altering a Hadamrd codebook (to break its equidistance property) 
and using a confusion matrix to greedily assign \emph{dissimilar} codewords to similar classes
(in contrast to our policy; see the discussion on this controversy above).

Both these works focus on specific greedy assignment policies for specific codebooks.
We on the other hand extensively test our hypotheses on many codebooks and  demonstrate the superiority of similarity-preserving assignments over similarity-breaking ones in traditional classification settings.
We exhaustively evaluate \emph{all} possible assignments in several codebooks on three small datasets (see \figref{fig:empirical_combined} and \appref{app:correlation});
and also evaluate different greedy assignment policies on larger datasets (see
\ref{sec:extreme_experiments} and
\appref{app:cifar100}).

%% file: supp.tex
\section{Error Correcting Codes: Illustration}
\label{app:basics}

To make our paper more approachable for readers who are less familiar with the Error-Correcting Codes scheme (\secref{sec:ecoc}), 
we now present a brief illustration of the entire scheme.
For further explanations, we recommend Section~3 \linebreak
in \citet{allwein2000loss}.

\begin{figure}[ht!]
\centering
\begin{minipage}{.9\textwidth}
  \vskip -.2cm
    \begin{subfigure}[b]{.317\linewidth}
      \centering
    \includegraphics[width=1\linewidth]{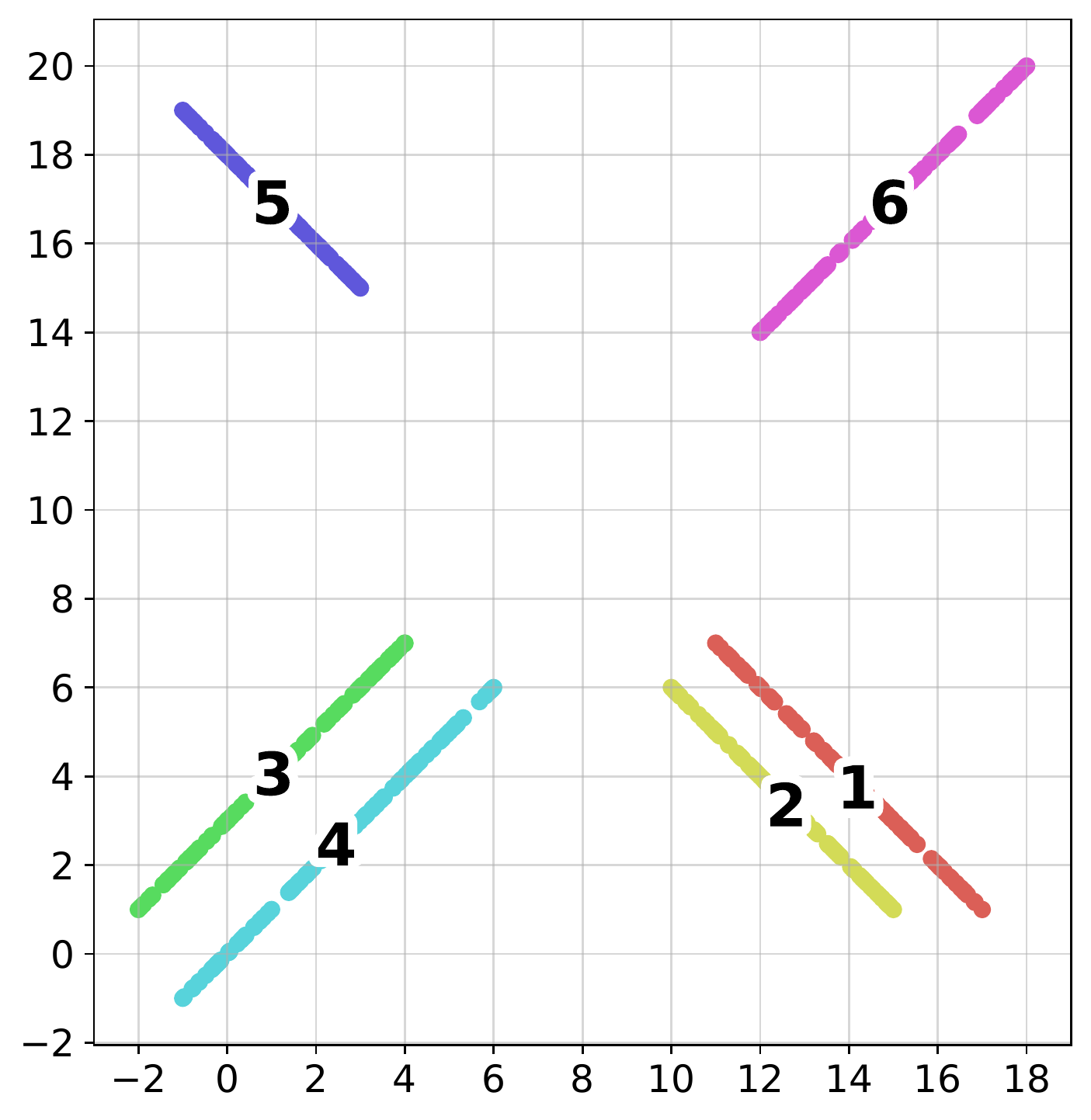}
    \vskip -.1cm
        \caption{The synthetic dataset
        \label{fig:synthetic}
        }
    \end{subfigure}
    \hspace{.005\textwidth}
    \begin{subfigure}[b]{.285\linewidth}
      \centering
    \includegraphics[width=1\linewidth]{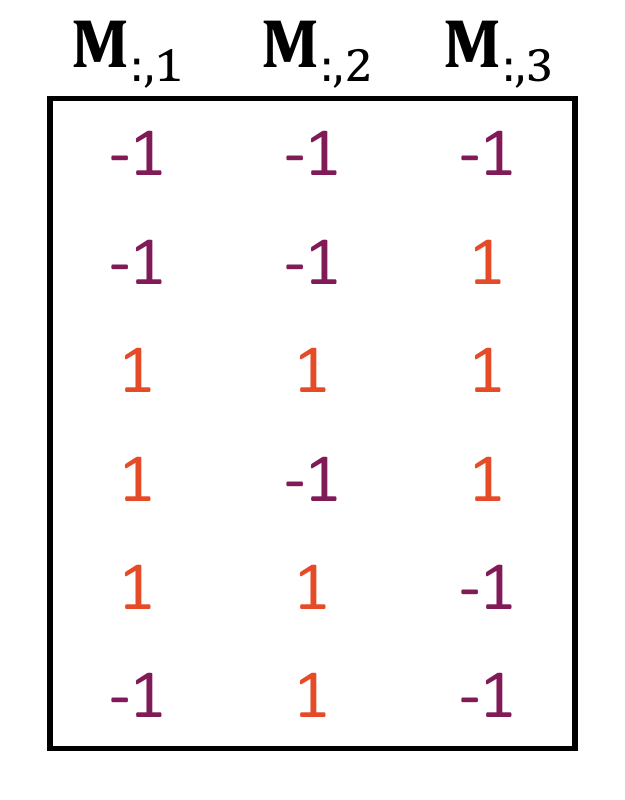}
    \vskip -.1cm
        \caption{A given codebook $\M$
        \label{fig:codebook}
        }
    \end{subfigure}
    %
    %\hspace{.005\textwidth}
    \begin{subfigure}[b]{.317\textwidth}
      \centering
      \includegraphics[width=1\linewidth]{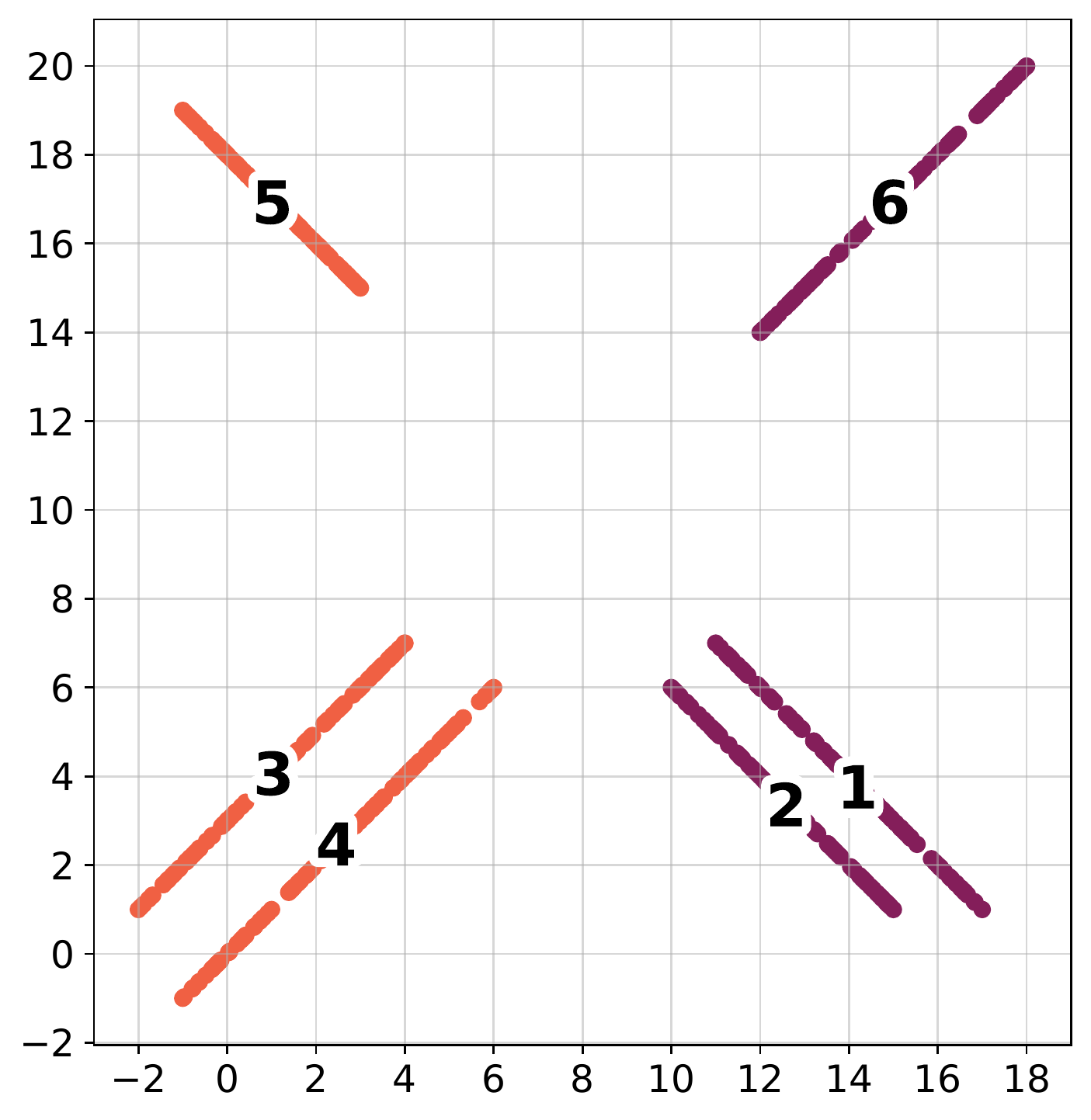}
    \vskip -.1cm
      \caption{Subproblem \#1
        \label{fig:subproblem}}
    \end{subfigure}
    \vskip -.1cm
    \caption{Given the synthetic dataset (left) and a given codebook (center), 
    assignments vary greatly in their accuracy (right).
    The best assignments achieve 100\% accuracy, while the worst achieve 37.17\%.}
    \label{fig:illustration}
\end{minipage}
\end{figure}

In this section and the next, we use a synthetic dataset with $K\!=\!6$ classes and $m=600$ training samples (100 per class).
The dataset is illustrated in \figref{fig:synthetic}.

For simplicity, 
in \figref{fig:codebook}
we present a small codebook with $\ell\!=\!3$ columns and no redundancy.
Each column of the codebook $\M$ induces a binary subproblem.
One such subproblem, corresponding to the leftmost column,
is depicted in \figref{fig:subproblem}
and requires separating classes $1,2,6$ from $3,4,5$.
Each binary subproblem is learned by a model of choice, \eg SVM or a decision tree, yielding $\ell$ binary predictors $f_1,\dots,f_\ell$.

%\vskip 2cm

At test time, given an input $\vx$, the binary predictors output a prediction vector
$\vf\!\left(\vx\right)
\triangleq
\left[f_1\!\left(\vx\right)\!, 
\smalldots,
f_\ell\!\left(\vx\right)\right]^\top
\!$.
In turn, the final prediction is made either by thresholding $\vf\!\left(\vx\right)$ and looking for the nearest neighbor (row) of the codebook $\M$, or by a more sophisticated decoding scheme that takes into account the prediction magnitudes as well (see \eqref{eq:soft-decoding}).
\linebreak
For instance, if 
$\vf\!\left(\vx\right)
=
\left[0.1, -3, 2.4\right]^\top
\!$, then a hard decoding scheme, which is equivalent to nearest-neighbor decoding, will compute 
$\sgn\left(\vf\!\left(\vx\right)\right)
=
\left[1,-1,1\right]^\top
\!$, and the prediction would be $\hat{y}(\vx)=4$
(see the fourth row in \figref{fig:codebook}).

\vfill

\newpage

\section{Synthetic dataset}
\label{app:synthetic}

Here we illustrate the importance of codeword-to-class assignments using the synthetic dataset from the previous section ($K\!=\!6$ classes, $m=600$ training samples).
For simplicity, we use a small codebook with $\ell\!=\!3$ columns and no redundancy.

\begin{figure}[ht!]
    \vskip -.3cm
    %\vskip 1cm
    \centering
\begin{minipage}{.9\textwidth}
    \begin{subfigure}[b]{.25\textwidth}
      \centering
      \includegraphics[width=1\linewidth]{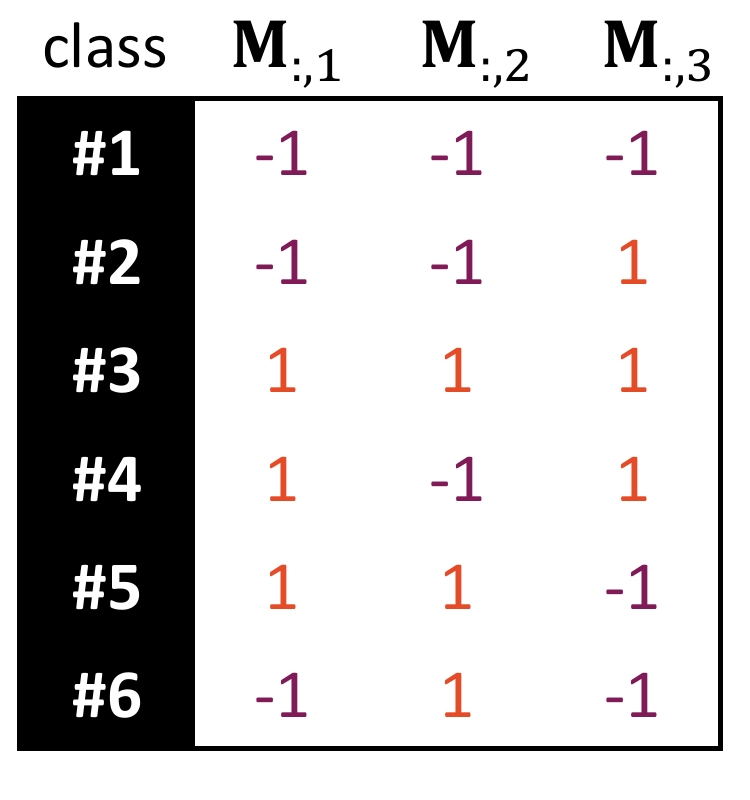}
    \vskip -.1cm
        \caption{Assigned codebook}
    \end{subfigure}
    \hspace{.005\textwidth}
    \begin{subfigure}[b]{.23\textwidth}
      \centering
      \includegraphics[width=1\linewidth]{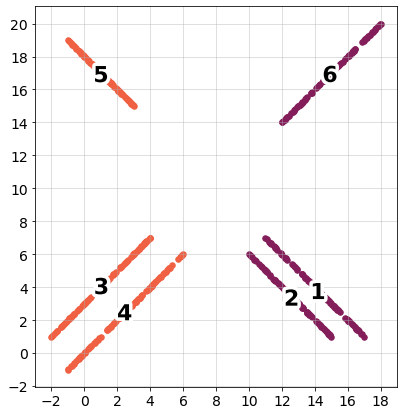}
    \vskip -.1cm
      \caption{Subproblem \#1}
    \end{subfigure}
    \hspace{.005\textwidth}
    \begin{subfigure}[b]{.23\textwidth}
      \centering
      \includegraphics[width=1\linewidth]{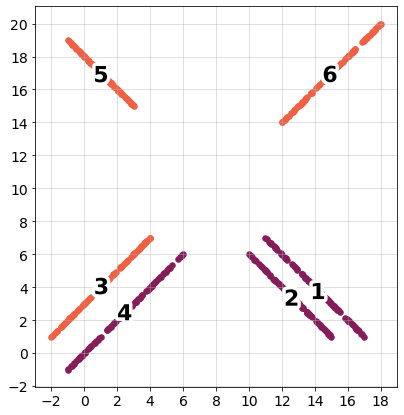}
    \vskip -.1cm
    \caption{Subproblem \#2
    }
    \end{subfigure}
    \hspace{.005\textwidth}
    \begin{subfigure}[b]{.23\textwidth}
      \centering
      \includegraphics[width=1\linewidth]{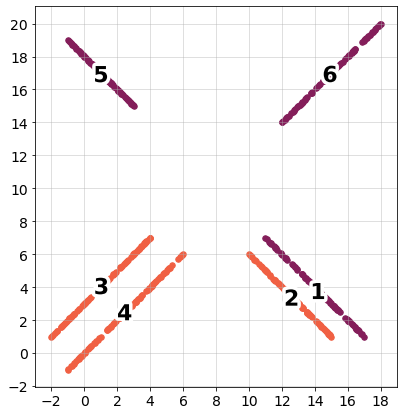}
    \vskip -.1cm
        \caption{Subproblem \#3}
    \end{subfigure}
    \vskip -.1cm
    \caption{Subproblems induced by the \emph{best} assignment (acc. = 100\%)
    are linearly \emph{separable}.}
    \label{fig:good_toy_assignment}
\end{minipage}
\end{figure}

For this codebook, only 12 out of 720 assignments can fit the data perfectly with a linear predictor.
These~assignments~all correspond to the same codebook (since the column order does not matter in ECC schemes and since complementary binary partitions are equivalent).
These assignments also beat one-vs-all (OVA) trained with a (tuned) linear SVM that achieves only 89.83\% (setting the Soft-SVM's $C$ as $0.89$).
Finally, the best assignments apparently preserve similarity (see \figref{fig:good_toy_assignment} and compare the codewords of the neighboring classes \#1 and \#2 to those of \#4 and \#6).

%\vskip 2cm

\begin{figure}[ht!]
\centering
\begin{minipage}{.9\textwidth}
  \vskip -.2cm
    \begin{subfigure}[b]{.317\linewidth}
      \centering
    \includegraphics[width=1\linewidth]{figs/toy.pdf}
    \vskip -.1cm
        \caption{The synthetic dataset}
    \end{subfigure}
    \hspace{.005\textwidth}
    \begin{subfigure}[b]{.285\linewidth}
      \centering
    \includegraphics[width=1\linewidth]{figs/toy_codebook.pdf}
    \vskip -.1cm
        \caption{A given codebook $\M$}
    \end{subfigure}
    %
    %\hspace{.005\textwidth}
    \begin{subfigure}[b]{.38\linewidth}
      \centering
    \includegraphics[width=1\linewidth]{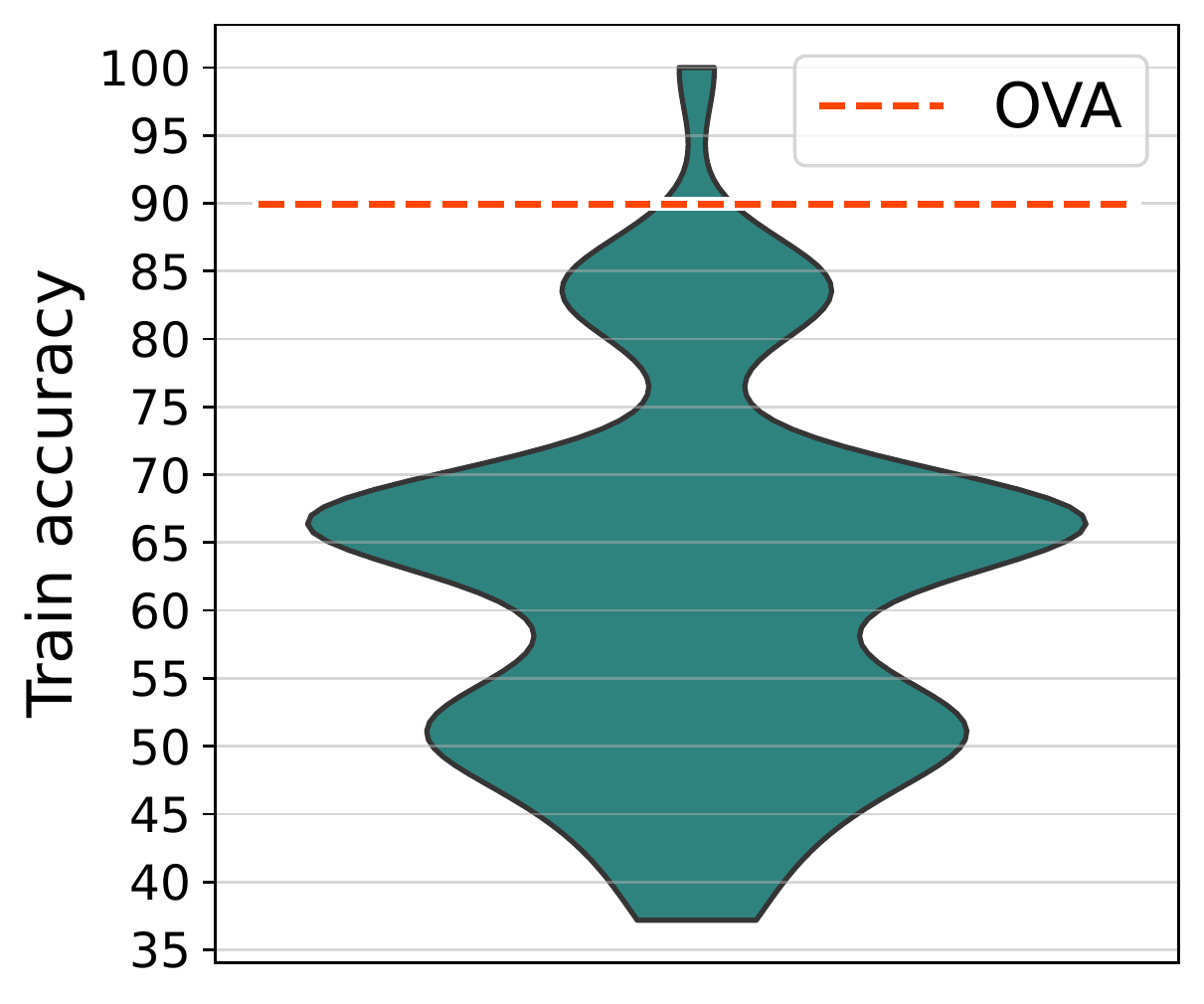}
    \vspace*{-2.5mm}
    \vskip -.1cm
    \caption{Accuracy variation of assignments}
    \end{subfigure}
    \vskip -.1cm
    \caption{Given the synthetic dataset (left) and a given codebook (center), 
    assignments vary greatly in their accuracy (right).
    The best assignments achieve 100\% accuracy, while the worst achieve 37.17\%.}
    % \label{fig:class_codeword}
\end{minipage}
\end{figure}

%\newpage

Now we illustrate \emph{why} the subproblems induced by the best assignment are inherently easier than the ones induced by the worst assignment 
(using the same codebook).

%\vskip 1cm

\begin{figure}[ht!]
  \vskip -.5cm
    \centering
\begin{minipage}{.9\textwidth}
    \begin{subfigure}[b]{.25\textwidth}
      \centering
      \includegraphics[width=1\linewidth]{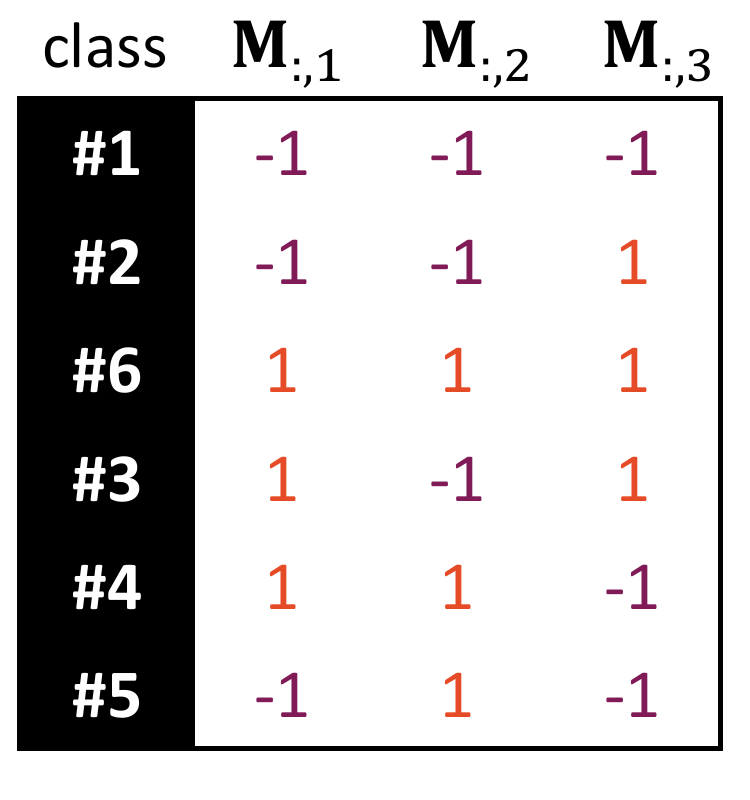}
    \vskip -.1cm
        \caption{Assigned codebook}
    \end{subfigure}
    \hspace{.005\textwidth}
    \begin{subfigure}[b]{.23\textwidth}
      \centering
      \includegraphics[width=1\linewidth]{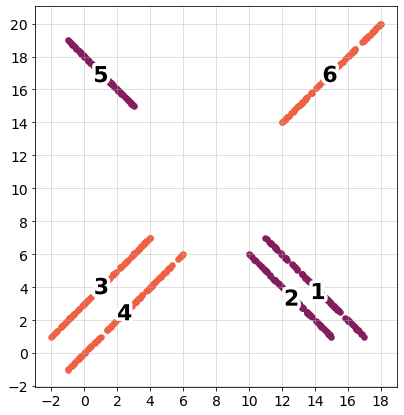}
    \vskip -.1cm
      \caption{Subproblem \#1}
    \end{subfigure}
    \hspace{.005\textwidth}
    \begin{subfigure}[b]{.23\textwidth}
      \centering
      \includegraphics[width=1\linewidth]{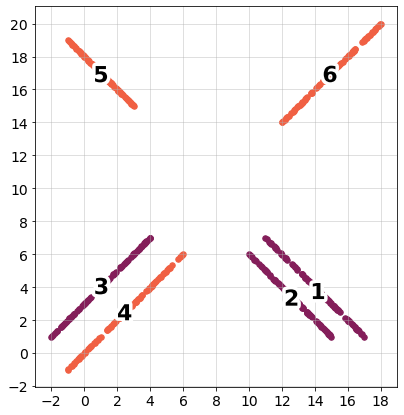}
    \vskip -.1cm
    \caption{Subproblem \#2
    }
    \end{subfigure}
    \hspace{.005\textwidth}
    \begin{subfigure}[b]{.23\textwidth}
      \centering
      \includegraphics[width=1\linewidth]{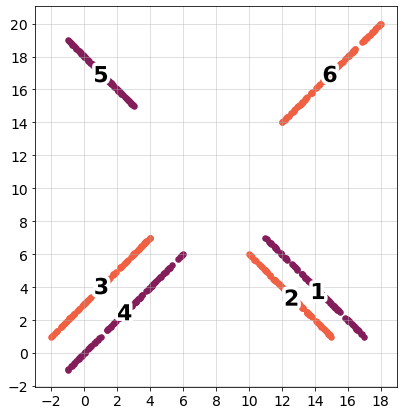}
    \vskip -.1cm
        \caption{Subproblem \#3}
    \end{subfigure}
    \vskip -.1cm
    \caption{Subproblems induced by the \emph{worst} assignment (acc. = 37.17\%) are linearly \emph{inseparable}.}
    % \label{fig:class_codeword}
\end{minipage}
\end{figure}

\newpage

\section{Training details for the exhaustive experiments in \secref{sec:exp-exhaustive}}
\label{app:training_exhaustive}

\subsection{Evaluating all possible assignments}
\label{app:creating_assignments}

The exhaustive experiments require obtaining the test accuracy of every possible codeword-to-class assignment of given codebooks.
Following are the training details of our experimental setup.
This exhaustive setup resembles of the setup in \citet{bai2016ecoc}, but we use it as a means for simply showing that similar codewords should be assigned to similar classes, while they propose it as a practical approach for small datasets (with very few classes).

The datasets used in the exhaustive experiments have $K=10$ classes each.
This means that each dataset has $K!\approx 3.6M$ assignments.
Instead of training every assignment from scratch, we notice that there are at most $2^K=1,024$ possible binary columns with $K=10$ rows.
We further notice that a column $\M_{:,j}$ and its complementary column $-\M_{:,j}$ create the same binary classification task (with opposite labels), thus reducing the number of possible binary partitions to $2^{K-1}=512$.
Finally, columns consisting of only $+1$ (or $-1$) induce meaningless partitions. Hence, the number of binary partitions we actually need to train on is $2^{K-1}-1=511$.

The aforementioned $511$ columns constitute every possible codebook with $K=10$ codewords.
Specifically, given a codebook, all possible assignments correspond to all possible row permutations of the codebook.
Thus, instead of training 3.6M codebooks, we train only 511 columns, construct every possible assignment (permuted codebook) from the pretrained columns, and finally, merely check the test accuracy of the resulting codebook.

Clearly, the trick above reduces the time required for our experiments (since training is much more expensive than inference).
Moreover, it reduces the variation in the test accuracies stemming from the training itself, since a binary partition that appears in multiple assignments is only trained once.
This makes the observed variation in \secref{sec:variability} more significant.

%\newpage

\bigskip

\subsection{Hyperparameter tuning}
\label{app:tuning}

We now explain how we train the $511$ binary partitions of each dataset.

\subsubsection{\mnist and \cifarA}
For these two datasets, we use the (soft-margin) SVM algorithm \citep{cortes1995svm} as the base learner.
The SVM problem is
\vspace{-0.2em}
\begin{align*}
 \min_{\pmb{w},b}
 \frac{1}{2} \norm{\pmb{w}}^2
 + C\sum_{i=1}^{m}
 \max\{0, \,1-y_i(\pmb{w}^\top \pmb{x}_i+b)\}
 \,,
\end{align*}
where $C>0$ is a regularization parameter that requires tuning.

To tune $C$, 
we perform $3$-fold cross-validation on the training sets (only),
evaluating the performance of
\linebreak
$C\in\{10^{-3}, 10^{-2}, \dots, 10^{2}, 10^{3}\}$.
Consequently, we choose $C=10$ for \mnist and $C=1$ for \cifarA.

\subsubsection{\yeast}

For this dataset, the base learners are decision trees with the Gini splitting criterion.
Nodes are split until they are pure or until they contain less than 3 examples.
The minimum sample-number for splitting (\ie 3) was chosen from $\left\{1,3,5,7\right\}$ after we found it yielded the best validation performance.

\newpage

\subsection{Building class metrics}
\label{app:class_metrics}

The class-codeword score \eqref{eq:cc_dist} described in \secref{sec:assignments_importance} requires a class metric in the form of a distance matrix 
$\Dcls\in\doubleR_{\ge0}^{K\times K}$.
In \secref{sec:exp_class_similarity} we construct the aforementioned class distance matrix in two ways:
(a) using a confusion matrix, and (b) using class means (of raw features).
Later in \appref{app:cifar100}, we also use word embeddings of class names.
Details follow.

\subsubsection{Using a confusion matrix}
\label{app:construct_confusion}

We use confusion matrices as a similarity measure on classes, assuming confusable classes are semantically similar.
For each dataset, we train a one-vs-all (OVA) classifier and compute its confusion matrix.
For \mnist and \cifarA, we compute the confusion matrix on the training set itself (since there are sufficient errors for the matrix to be informative).
For \yeast, there are very few training samples and the models have a high capacity (decision trees), resulting in very few training errors and a sparse confusion matrix.
Thus, we split the training set of \yeast into 8 folds. We train on one fold and compute the confusion matrix on the other 7 folds. Finally, we sum the 8 resulting confusion matrices.

Each confusion matrix $\Ccls$ is an asymmetric similarity matrix (where the sum of all entries is 1), while we require $\Dcls$ to be a symmetric distance matrix.
At first, we considered using
\begin{align}
    \A \triangleq
   \pmb{1}_{K\times K} - \frac{1}{2}\left(\Ccls + \Ccls^\top \right)~,
\end{align}
which is a symmetric dissimilarity measure as required. 
However, since the entries of $\Ccls$ are often very close to $0$, the above transformation yields a matrix that is very close to a rank-one matrix which is substantially different from $\Dcw$ (the eigenvalue analysis in \citep{umeyama1988WGMP} shows why this is problematic).

To induce reasonable spectra, we used the following matrix (simpler matrices yield mostly similar results in our experiments):
\begin{align}
    \begin{split}
    \forall i: D_{i,i} & = 0,
    \\
    \forall i\neq j:
    D_{i,j} 
    &=
    \log \left( A_{i,j} \right)
    -
    1.1
    \min_{i'\neq j'}{
    \log \left( A_{i',j'} \right)}
    ~,
    \end{split}
\end{align}
which yields a symmetric dissimilarity matrix, with a vast spectrum of eigenvalues, as depicted below.
In practice, the $\Dcls$ matrices stemming from the above approach are very informative and create valuable class-codeword scores which are highly correlated with the test accuracy (see \secref{sec:exp_class_similarity}).

\vspace{1.3em}

\begin{figure}[ht!]
\centering
  \hfill
  \begin{minipage}[t!]{0.35\linewidth}
    \caption{
    The class metric matrix $\Dcls$ of \mnist, 
    constructed from the confusion matrix as described above.
    Notice how this matrix quantifies visual class semantics (similarities).
    For instance, see how $4$ has a small distance from $9$ and how $3$ is close to $2,5,8$, etc.
    \label{fig:confusion}
    }
  \end{minipage}
  \hfill
\begin{minipage}{.55\textwidth}
    \centering
    \includegraphics[width=.9\linewidth]
      {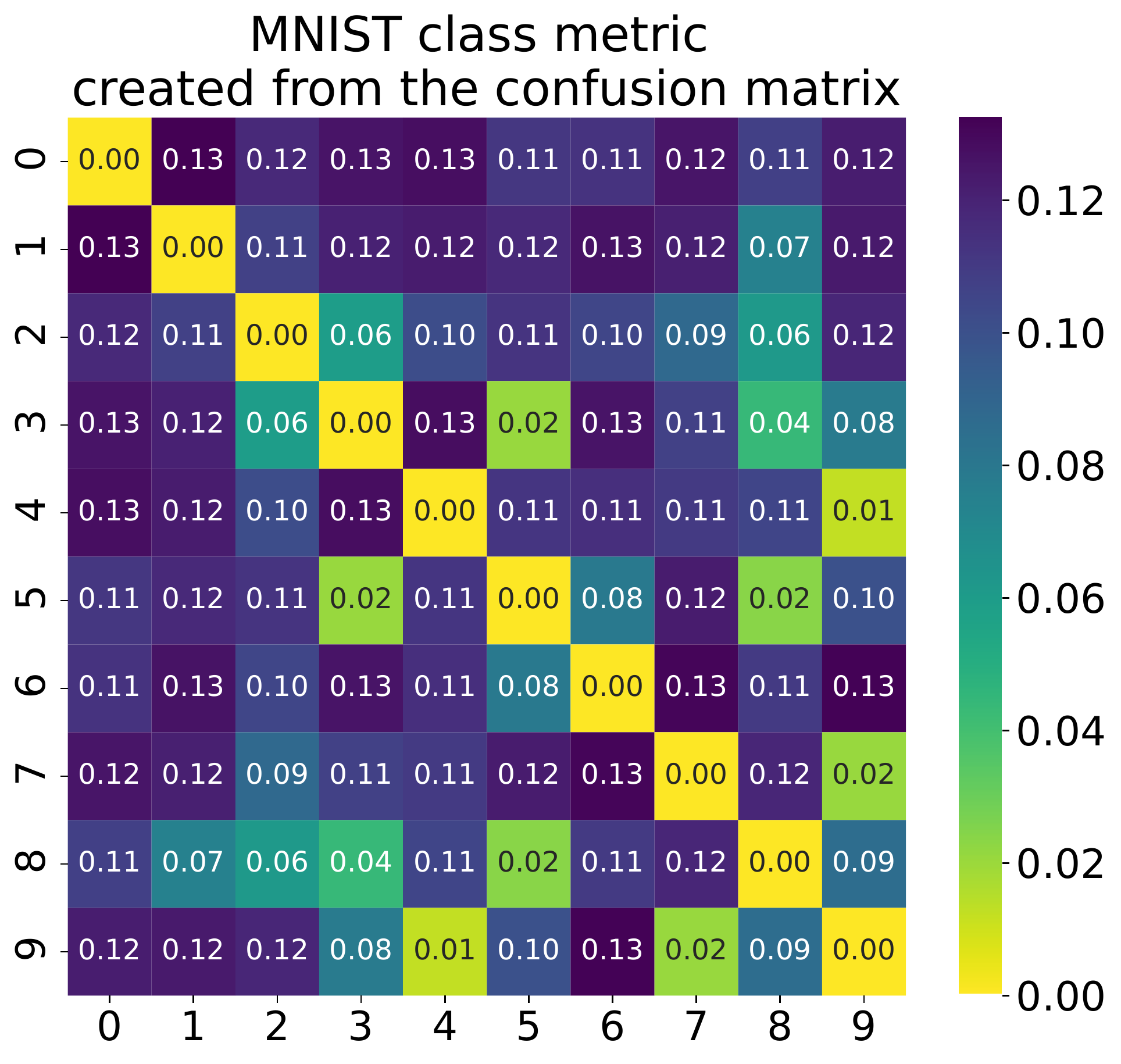}
\end{minipage}
  \hfill
\end{figure}

\newpage

\subsubsection{Using class means of raw features}
\label{app:construct_means}

For each dataset, we embed each class in a high-dimensional Euclidean space by computing the means of the \emph{raw} features of all \emph{training} samples of that class.
Then, we set $\Dcls$ as the matrix of Euclidean distances between these embeddings.
The resulting matrices are symmetric and the class-codeword scores stemming from them are informative, as seen from the apparent correlations to the test accuracy
(see \appref{app:correlation}).

Importantly, these embeddings do not require actually training other models (unlike confusion matrices).
Moreover, since we use simple base learner (\eg linear models), one should expect that similarity-preserving assignments according to the proposed $\Dcls$ will ``concentrate'' the samples of each binary class in each of the induced binary subproblems in the Euclidean space, thus creating easier subproblems.

\vspace{1em}

\begin{figure}[ht!]
\centering
  \hfill
  \begin{minipage}[t!]{0.35\linewidth}
    \caption{
    The class metric matrix $\Dcls$ of \mnist, 
    constructed from the class raw feature means as described above.
    Notice how this matrix quantifies visual class semantics (similarities) as well, and how it resembles the metric created from the confusion matrix (\figref{fig:confusion}).
    }
  \end{minipage}
  \hfill
\begin{minipage}{.55\textwidth}
    \centering
    \includegraphics[width=.9\linewidth]
      {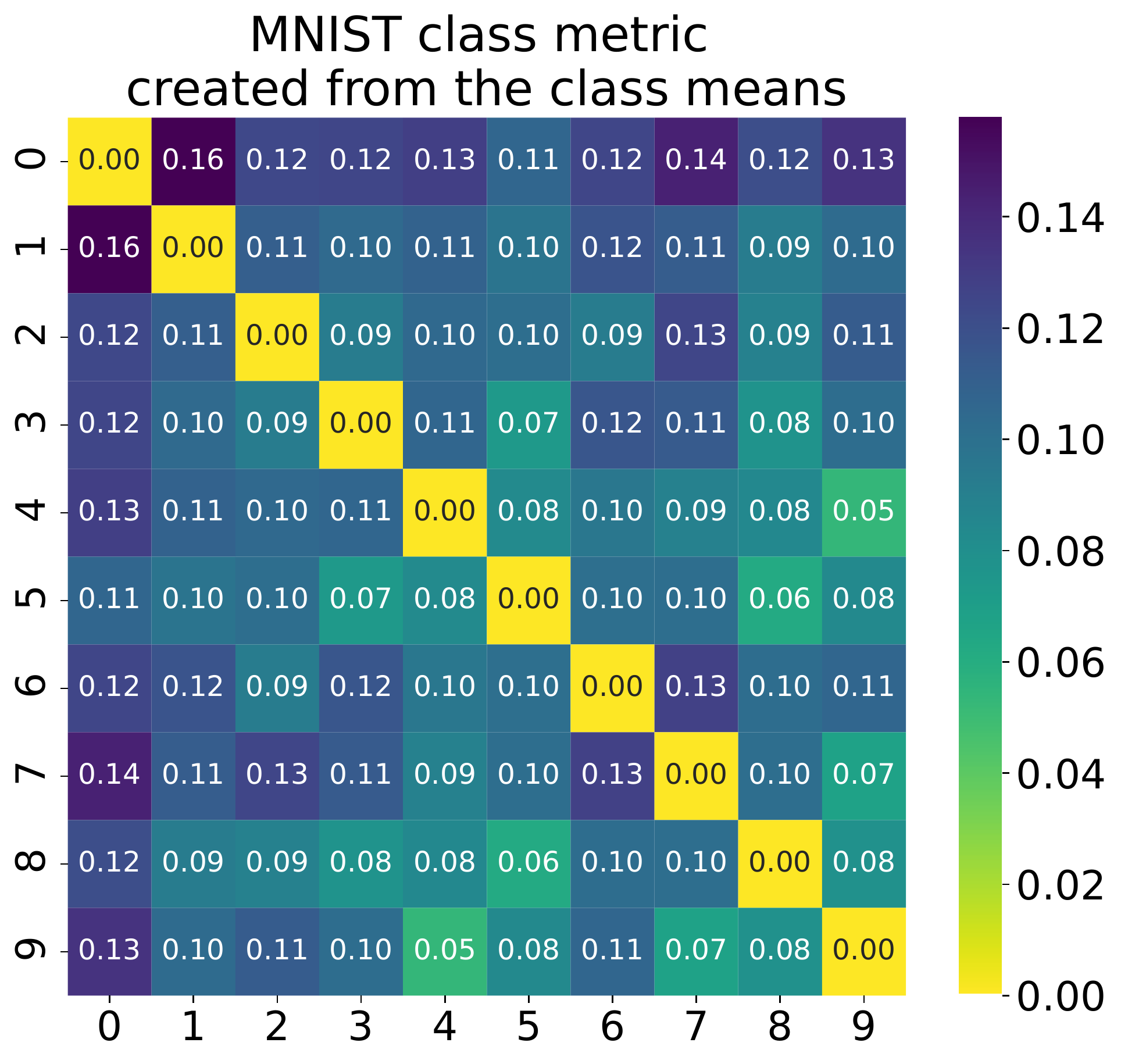}
\end{minipage}
  \hfill
\end{figure}

\newpage

\section{All correlation graphs for the exhaustive experiments in \secref{sec:exp-exhaustive}}
\label{app:correlation}

Now we show similar results and correlations to the ones shown in \secref{sec:exp-exhaustive} for additional two codebooks ---
\linebreak
Random dense $10\!\times\! 15$ and Spectral $10 \!\times\! 8$ 
(built using the method from \citealp{zhang2009spectral}).
Moreover, we also report results using class-codeword scores stemming from the means of the raw features of each class rather than confusion matrices (see \appref{app:construct_means}).

Our observations and findings evidently apply to many codebooks and class metrics 
(see \appref{app:cifar100} for an additional metric).

\paragraph{How to understand the plots?}
Each level set contains $\approx\! 10\%$ of all possible 3.6M assignments.
    The $10^{-3}$ least probable assignments are scattered
    as individual points.
    Regressors computed on all assignments are plotted in orange.
    Also written are the coefficients of determination ($r^2$).
    
\subsection{\mnist}

In \mnist, the average binary loss $\varepsilon$ is highly correlated with the test accuracy for the three tested codebooks.
Our method for constructing $\Dcls$ from confusion matrices (described in \appref{app:construct_confusion}) yields class-codeword scores that are correlated to the test accuracy (but less than the average binary loss).
Finally, the class-codeword scores stemming from the means of the raw features (see \appref{app:construct_means}) 
are the least correlated to the test accuracy, but are still informative.

\begin{figure}[ht!]
    \vskip .2cm
    \centering
  \centering
  \includegraphics[width=.98\linewidth]{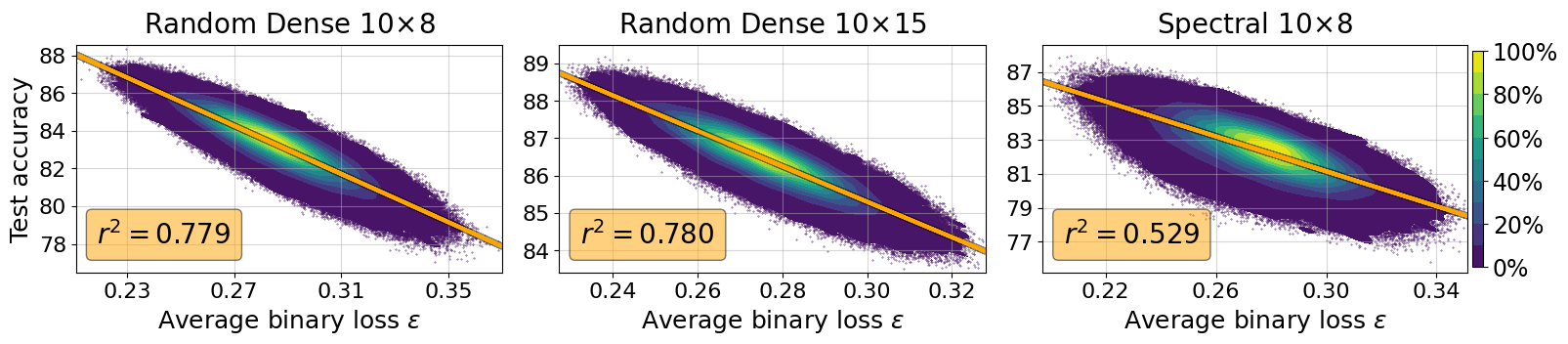}
  \vskip -2mm
    \caption{Test accuracy vs. Average binary loss
    in
    \mnist.}
    % \label{fig:class_codeword}
\end{figure}

\begin{figure}[ht!]
    \vskip .2cm
    \centering
  \centering
  \includegraphics[width=.98\linewidth]{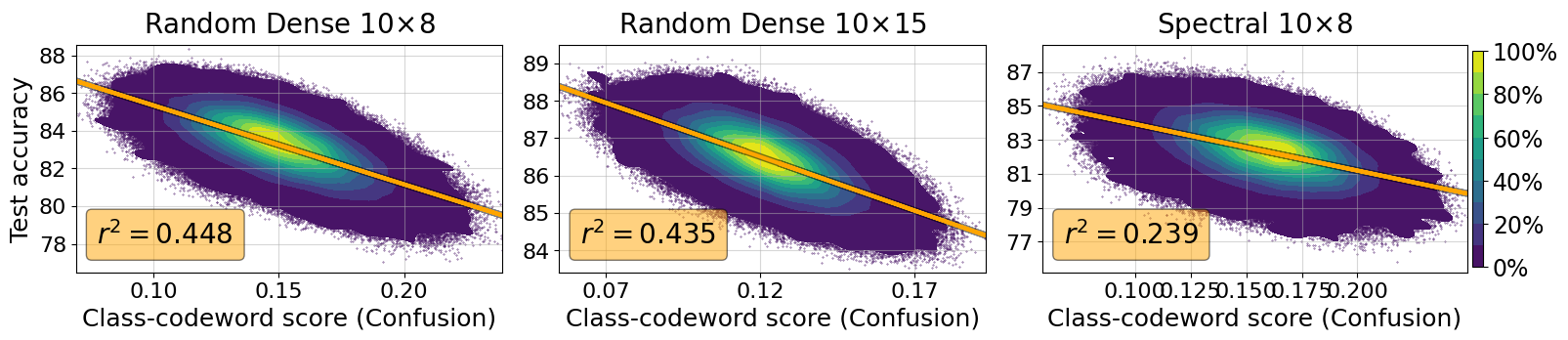}
  \vskip -2mm
    \caption{Test accuracy vs. Class-codeword scores 
    in
    \mnist, 
    using the confusion matrix to construct $\Dcls$.}
\end{figure}

\begin{figure}[ht!]
    \vskip .2cm
    \centering
  \centering
  \includegraphics[width=.98\linewidth]{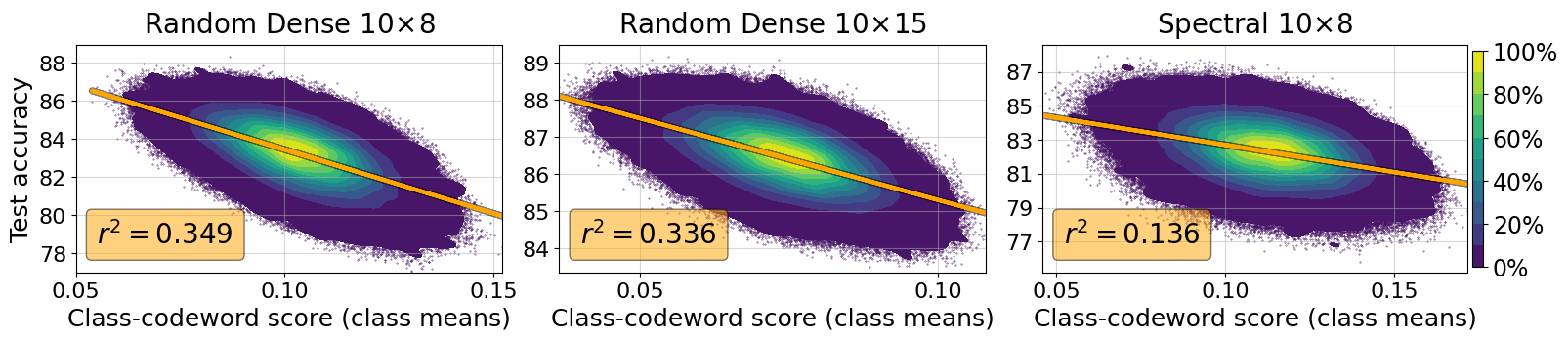}
  \vskip -2mm
    \caption{Test accuracy vs. Class-codeword scores 
    in
    \mnist, 
    using class (feature) means to construct $\Dcls$.}
\end{figure}

\newpage

\subsection{\cifarA}

In \cifarA, the test accuracy is less correlated to the average binary loss compared to the correlation in \mnist.
\linebreak
However, the class-codeword scores computed using the confusion matrices are very informative (sometimes even comparable with the average binary loss).
The class means are again slightly less informative than the confusion matrices.

\begin{figure}[ht!]
    \vskip 1.2cm
    \centering
  \centering
  \includegraphics[width=.98\linewidth]{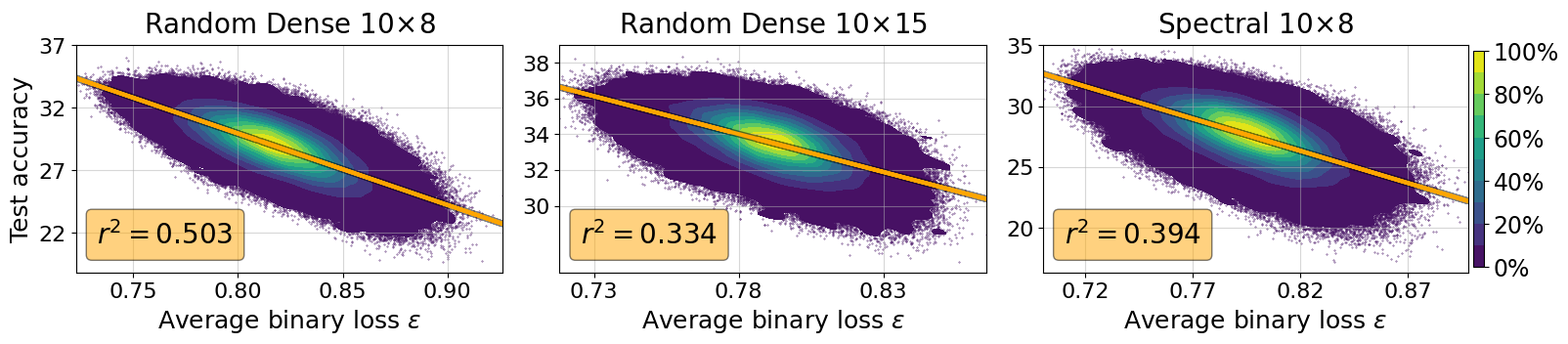}
  \vskip -2mm
    \caption{Test accuracy vs. Average binary loss
    in
    \cifarA.}
\end{figure}

\begin{figure}[ht!]
    \vskip 1.2cm
    \centering
  \centering
  \includegraphics[width=.98\linewidth]{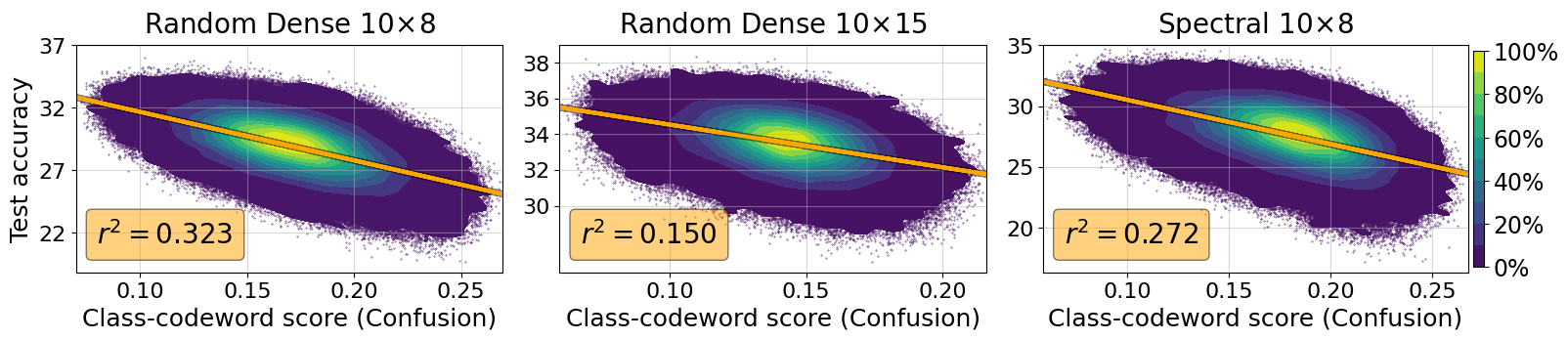}
  \vskip -2mm
    \caption{Test accuracy vs. Class-codeword scores 
    in
    \cifarA, 
    using the confusion matrix to construct $\Dcls$.}
\end{figure}

\begin{figure}[ht!]
    \vskip 1.2cm
    \centering
  \centering
  \includegraphics[width=.98\linewidth]{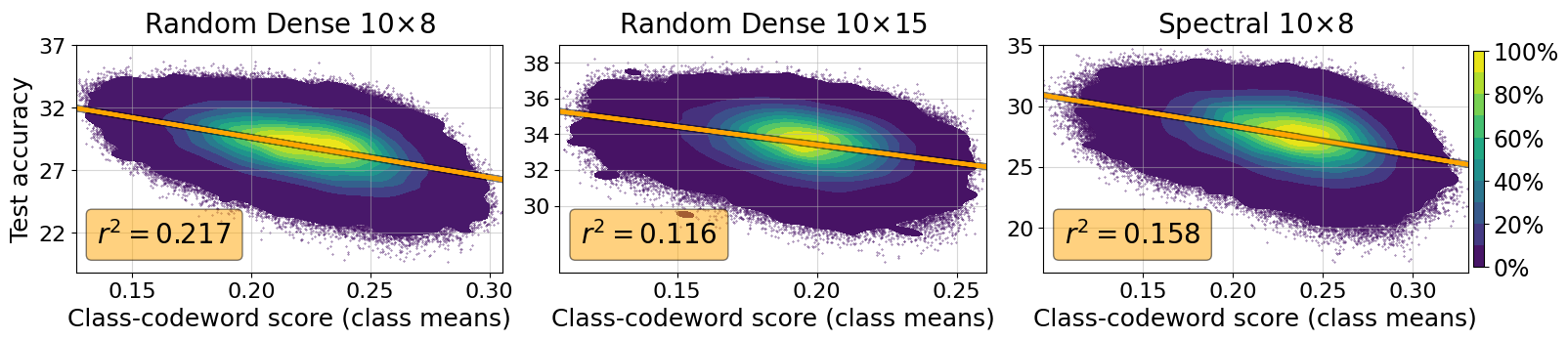}
  \vskip -2mm
    \caption{Test accuracy vs. Class-codeword scores 
    in
    \cifarA, 
    using class (feature) means to construct $\Dcls$.}
\end{figure}

\newpage

\subsection{\yeast}
\yeast exhibits the worst correlations among the three datasets, but the assignments still evidently vary, and their accuracy is mildly controlled by the class-codeword scores.
One should also notice that this dataset is much smaller than the other two (in both the number of training examples and number of features, see \tabref{tbl:datasets}),
which might explain the larger variation and lower correlations it exhibits.

\begin{figure}[ht!]
    \vskip 1.2cm
    \centering
  \centering
  \includegraphics[width=.98\linewidth]{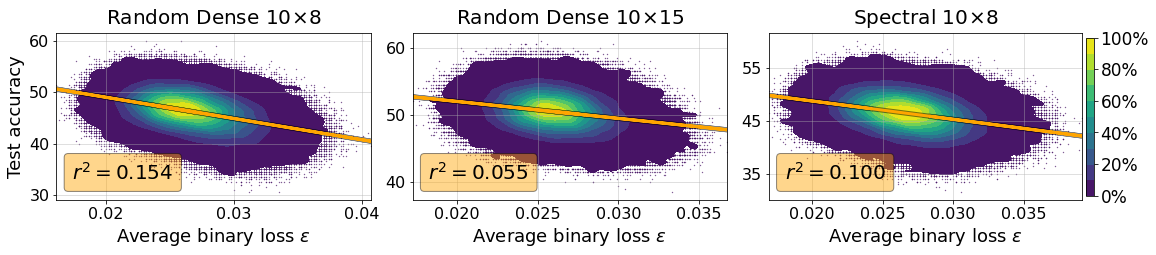}
  \vskip -2mm
    \caption{Test accuracy vs. Average binary loss
    in
    \yeast.}
\end{figure}

\begin{figure}[ht!]
    \vskip 1.2cm
    \centering
  \centering
  \includegraphics[width=.98\linewidth]{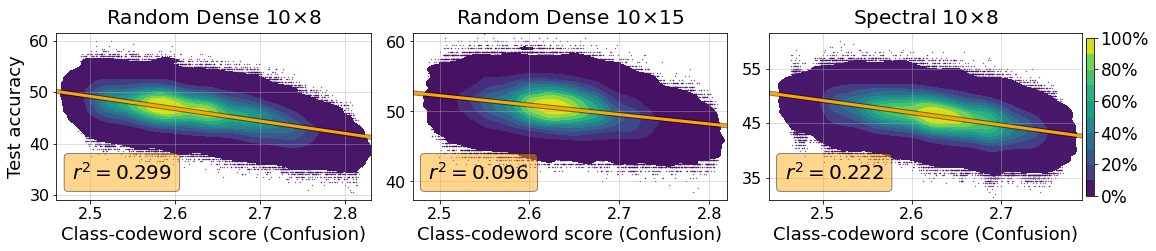}
  \vskip -2mm
    \caption{Test accuracy vs. Class-codeword scores 
    in
    \yeast, 
    using the confusion matrix to construct $\Dcls$.}
\end{figure}

\newpage

\section{Intermediate scale: \cifarB}
\label{app:cifar100}

In this section, we use \cifarB \citep{krizhevsky2009cifar} that have $K=100$ classes to demonstrate that the class-codeword score is correlated with the test accuracy in a larger dataset than the ones we use in \secref{sec:exp-exhaustive}.
Moreover, we show that as we discuss in \secref{sec:related}, the class-codeword score \eqref{eq:cc_dist} can also be used to efficiently \emph{find} a similarity-preserving assignment in practice.
So far in our other experiments in \secref{sec:experiments}, 
we did not explicitly use the score to find good assignments.
In the small-scale experiments in \secref{sec:exp-exhaustive}, we exhaustively computed the test accuracy of all $K!$ possible assignments to empirically \emph{prove} the correlation to the class-codeword score.
In the extreme experiments in \secref{sec:extreme_experiments}, we employed a special structure of the \wltls codebooks and the class taxonomy of the extreme datasets we used
(but the class-codeword score was not \emph{explicitly} minimized).
Here on the other hand, given a codebook and a general class metric (not in a tree structure),
we explicitly minimize the class-codeword score using a local search algorithm to obtain a similarity-preserving assignment.

\begin{figure}[ht!]
\centering
\begin{minipage}{.8\textwidth}
    \centering
    \includegraphics[width=.6\linewidth]
      {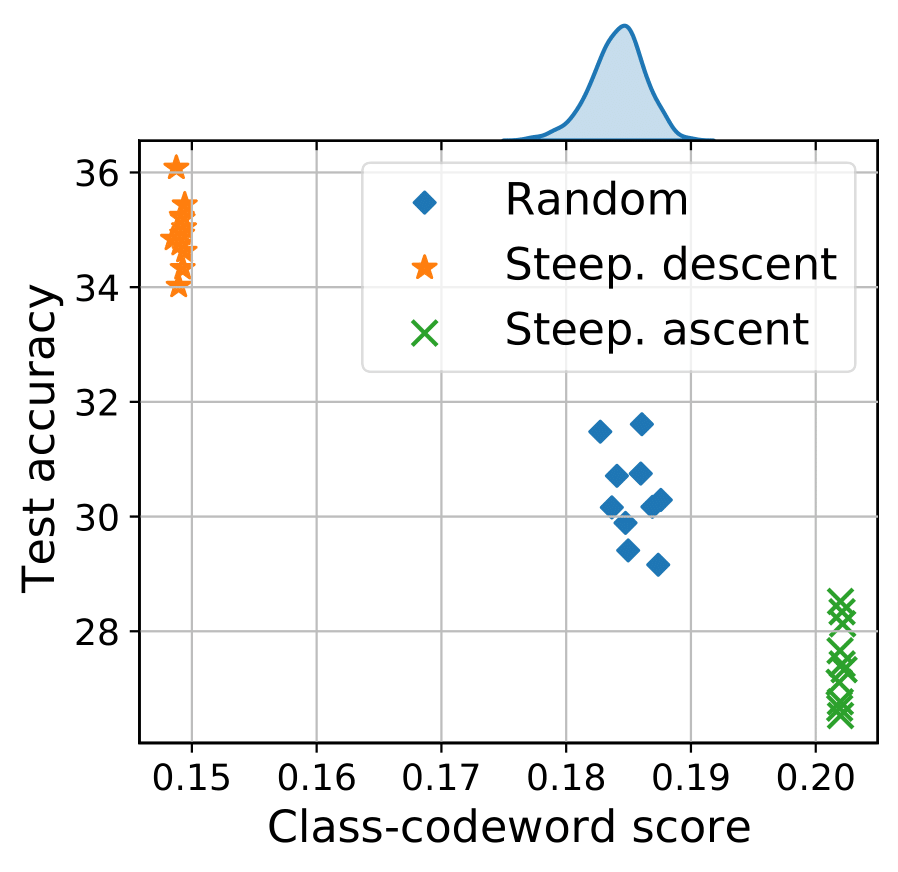}
    \caption{Results on a {\tt Dense} $10\times20$ codebook for \cifarB.
      On top is the empirical marginal distribution of class-codeword scores of random assignments.
    The class-codeword score is clearly correlated with the test accuracy.
    The local search algorithm finds similarity-preserving assignments that significantly improve on random assignments.
    Further discussion below.}
    \label{fig:cifar100-specific}
\end{minipage}
\end{figure}

\paragraph{Codebook.}
For this experiment, we use a $100\times 20$
random dense codebook.
Like \cite{allwein2000loss}, we choose the random codebook by randomizing $10^4$ dense codebooks and taking the codebook with the largest minimal Hamming distance $\rho$.

\paragraph{Feature extraction.}
We use a simple publicly available pretrained convolutional neural network\footnote{\url{https://github.com/aaron-xichen/pytorch-playground/}} for feature extraction.
This allows us to train, like before, simple linear predictors on top of the extracted features.

\paragraph{Class metric.}
As a metric between classes, we use the Euclidean distances between 
word embeddings of class names.
Specifically, we use a publicly available
\texttt{fastText} \citep{bojanowski2016fasttext} model, pretrained on \texttt{Common Crawl} and \texttt{Wikipedia}.
These embeddings serve as a cheap heuristic of a semantic metric between classes,
which can possibly approximate the visual class similarity.
Class names' embeddings were previously used for computer vision tasks (\eg for few-shot learning; \citealp{Xing2019fewshot}).
We use this semantic metric since it can very easily be acquired in many real-world scenarios where classes have known names.
Note that any other metric between classes should work here since this is already the third class-(dis)similarity measure we explore in the paper
(together with confusion matrices and class means; see \appref{app:class_metrics}).

\pagebreak

\paragraph{Optimizing the class-codeword score (Steepest descent hill-``climbing'' local search).}
Given a codebook and a dataset,
finding a class assignment (out of $100!\approx 9.33\cdot 10^{157}$
assignments) with a low class-codeword score is a hard task (see discussion in \secref{sec:related}).
However, we are able to find assignments with a low class-codeword score by performing a simple (discrete)  steepest descent algorithm.\footnote{Start from a random assignment. Search all ${\binom{K}{2}}=4,950$ assignments obtained by swapping 
the codewords of two classes only. 
Pick the assignment with the lowest class-codeword score and repeat until convergence.}
This allows us to quickly find many assignments that are more than $10$ standard deviations (!) farther from the mean class-codeword score.

\bigskip

\paragraph{Comparing different assignments.}

We train the scheme on the following codeword-to-class assignments:
\begin{enumerate}[leftmargin=1.5em]
  \item $10$ random assignments;
  \item $10$ similarity-preserving assignments 
  (\ie having a low class-codeword score), 
  found by using $10$ random restarts of steepest descent;
  \item $10$ similarity-breaking assignments
  (\ie having a high class-codeword score), 
  found similarly using steepest \emph{ascent}.
\end{enumerate}
All $30$ assignments are learned separately, 
and their test accuracy is plotted in the above \figref{fig:cifar100-specific} against their class-codeword score.

\bigskip

\paragraph{Discussing \figref{fig:cifar100-specific}.}
The figure demonstrates how similarity-preserving assignments significantly improve the performance of a predefined codebook on intermediate scales of $K$ as well.
The distribution on top of the plot is the empirical marginal distribution of class-codeword scores of random assignments. 
Note that the assignments found by the local search algorithms could not have been found by simply sampling assignments.

We obviously cannot run an exhaustive search on the entire $100!$ possible assignments in order to test the correlation, but these $30$ assignments agree with our empirical findings from \secref{sec:exp_class_similarity} that a lower class-codeword score implies better multiclass performance.

\newpage

\section{Supplementary material for the extreme classification (XMC) experiments in \secref{sec:extreme_experiments}}
\label{app:xmc}

\subsection{Extended dataset descriptions}

The dataset properties are brought in the table below.

\begin{table}[ht!]
  \centering
  \caption{
  Extreme Benchmarks' Extended Properties.
  }
  %\vskip 1mm
  \begin{tabular}{l|l|ll|lll|ll|l}
    %\toprule 
    \multicolumn{2}{c}{}
    &
    \multicolumn{2}{c}{}
    & 
    \multicolumn{3}{c}{Split Details}
    & 
    \multicolumn{2}{c}{\wltls Arguments}
    & 
    \\
    \toprule
    Dataset & Area & Classes & Features
    & Train & Val. & Test
    & Epochs
    & Early Stop.
    & Similarity
    \\
    \midrule
    %\\
    {\tt{aloi}} %\citep{rocha2013multiclass,yen2016pdsparse}
    & Vision & 
    1K &
    637K
    &
    90K & 10K & 8K
    &
    8
    &
    Yes
    &
    Clustering
    \\
    \lshtc %\citep{Partalas2015LSHTC}
    & Text &
    12K
    & 1.2M
    &
    83.8K & 5K & 5K
    &
    5
    &
    Yes
    &
    Given
    \\
    \dmoz %\citep{Partalas2015LSHTC}
    & Text & 
    27K & 
    575K 
    &
    330K & 15K & 39.2K
    &
    3
    &
    Yes
    &
    Given
    \\
    \ODP %\citep{bennett2009refined}
    & Text & 
    104K & 
    423K &
    867K & - & 493K
    &
    5 &
    No
    &
    Clustering
    \\
    \bottomrule
  \end{tabular}
\end{table}

Now we elaborate on these benchmarks for the sake of completeness and reproducibility.

\paragraph{aloi.bin \citep{rocha2013multiclass}.}
Downloaded from the PD-Sparse \citep{yen2016pdsparse} repository.\footnote{\url{https://github.com/a061105/ExtremeMulticlass} \label{foot:pdsparse}}
The dataset was created by applying Random Binning Features \citep{rahimi2007random} on the images of the original {\tt aloi} dataset. 
See \citet{yen2016pdsparse} for more details.

\paragraph{\lshtc \citep{Partalas2015LSHTC}.}
Also called {\tt LSHTC2010} or {\tt Dmoz2010}.
Downloaded from the PD-Sparse repository.\text{\textsuperscript{\ref{foot:pdsparse}}}

\paragraph{\dmoz \citep{Partalas2015LSHTC}.}
Also called {\tt LSHTC2011} or {\tt Dmoz2011}.
Originally this is not a multi-class dataset but a multi-label dataset. However, only 11,121 out of 394,756 training samples have more than one label. We thus remove these samples and randomly split the remaining 384K samples into train, validation, and test sets.
A similar process was used to create the more common XMC dataset {\tt Dmoz} (used for example in \citealp{yen2016pdsparse,evron2018wltls}).
However, the leaves of \dmoz were merged to create {\tt Dmoz}, and so it has only 12K labels instead of 27K like in the dataset we use.

\paragraph{\ODP \citep{bennett2009refined}.}
Downloaded from the Vowpal Wabbit repository.\footnote{\url{https://github.com/VowpalWabbit/vowpal_wabbit/tree/master/demo/recall_tree}}
Only 867K out of 1.08M training samples and 394K out of 493K test samples are non-empty.
We remove the empty training samples but keep the empty test samples.

\newpage

\subsection{Additional algorithmic details for the naive assignment algorithm}
\label{app:extreme_alg_details}

We will now illustrate more thoroughly the assignment algorithm described in \secref{sec:extreme_experiments}
and how it yields similarity-preserving assignments.

\paragraph{Understanding the coding graphs of \wltls.}
On the left side of the figure, we illustrate the coding graph used in \wltls \citep{evron2018wltls} for $K\!=\!8$ classes.
The graph induces an error-correcting codebook as follows:
\begin{enumerate}[leftmargin=5mm,topsep=1pt,]\itemsep.1pt
\item
Each source-sink path in the graph corresponds to one class (notice that there are $8$ such paths).
\item
Each edge in the graph corresponds to one column (\ie bit) in the codebook. 
That is, each edge induces one binary subproblem, 
separating the paths (classes) that use this edge from the ones that do not.
\end{enumerate}
Overall, we see in the center of the figure that each path
corresponds to a codeword whose $+\!1$ bits correspond to the edges used in the path.

\begin{figure}[ht!]
\centering
\begin{minipage}{.99\textwidth}
    \centering
    \includegraphics[width=.99\linewidth]
      {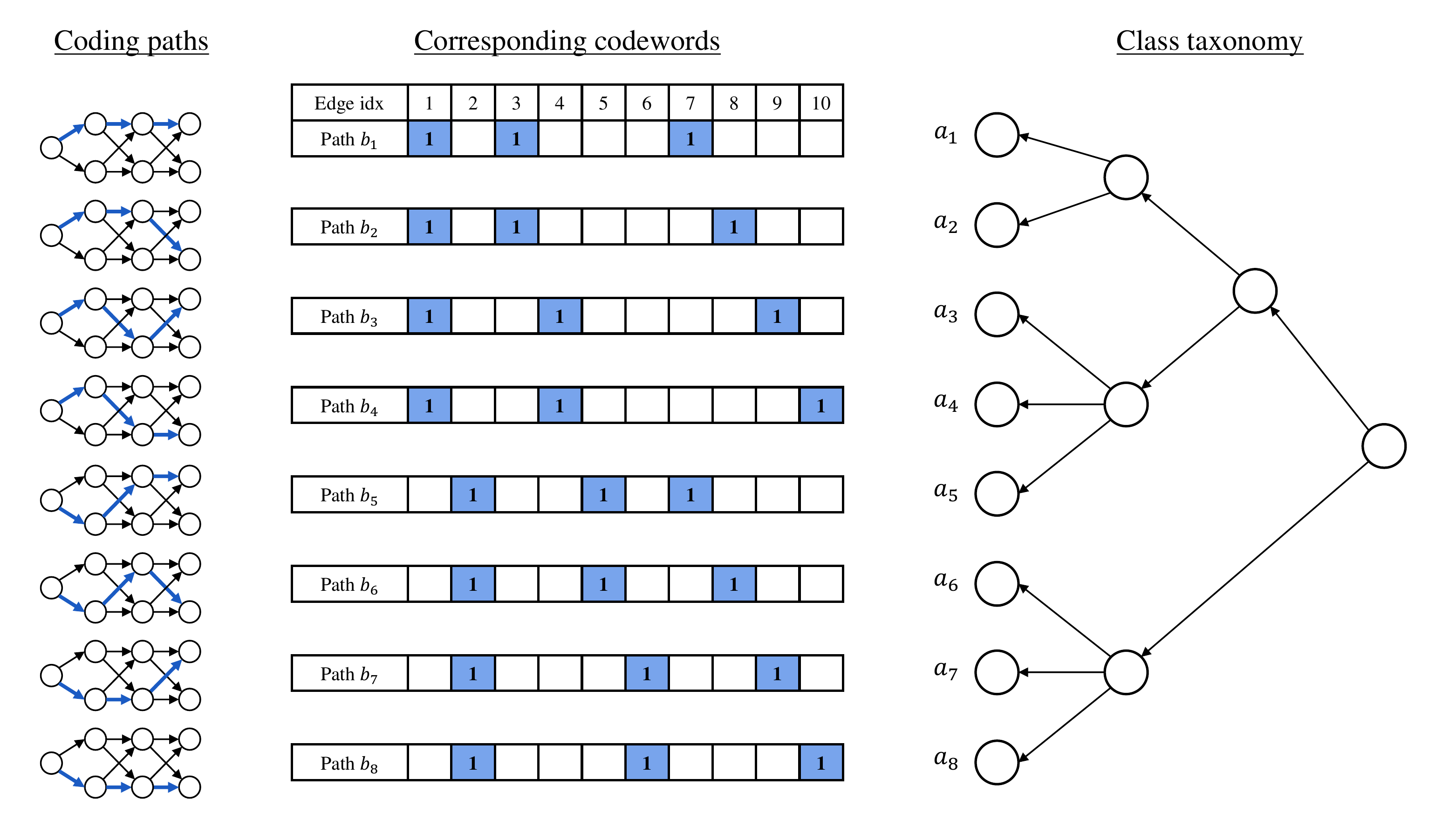}
\end{minipage}
\end{figure}

\paragraph{Understanding the assignment algorithm.}
We explain the algorithm according to its original steps:
\begin{enumerate}[leftmargin=5mm,topsep=1pt,]\itemsep.1pt
\item
Given some class taxonomy (either known-in-advance or computed by hierarchical clustering for instance),
the algorithm traverses the taxonomy to obtain an ordering $\left(a_1,\smalldots,a_K\right)$ of leaves (classes).
Notice that in most cases, classes $a_i$ and $a_{i+1}$ should be close on the taxonomy (\ie, these classes should be similar).
See the right side of the figure.
\item
Then, the algorithm recursively traverses the coding path (starting from the source; each node is visited \emph{many} times) to obtain an ordering $\left(b_1,\smalldots,b_K\right)$ of paths.
Due to the recursion, in most cases, paths $b_i$ and $b_{i+1}$ should be similar in edges (\ie, their codewords should be close).
See the left and center sides of the figure.
\item
Finally, since classes $a_i$ and $a_{i+1}$ should be similar 
and so should codewords $b_i$ and $b_{i+1}$,
then by assigning class $a_i$ to codeword $b_i$,
we create a similarity-preserving assignment.
\end{enumerate}

Like we explain in \secref{sec:extreme_experiments},
despite its simplicity, the illustrated algorithm succeeds in creating similarity-preserving assignments.
For instance, in \lshtc with $\ell=56$ edges, the algorithm found an assignment whose class-codeword score improves over the average score of random assignments by more than $900$ standard deviations.
Importantly, in all datasets, the multiclass accuracy significantly improves by using assignments found by the algorithm.

\newpage

\subsection{Tabular results}
\label{app:tabular}

Here we summarize the extreme classification experiments of \secref{sec:extreme_experiments}.
For both datasets, similarity-preserving assignments consistently and significantly beat the random assignments.
For all datasets, the accuracies are averaged over $5$ runs.
Two empirical standard deviations (of the runs, not their means) are reported as well.

\subsubsection{\aloi results of \figref{fig:aloi}}

\begin{table}[ht]
  %\vskip -.2cm
  \centering
  \begin{tabular}{llllll}
    \toprule
    Assignment method &
    $\ell=42$ ($b=2$) & 
    $\ell=55$ ($b=3$) & 
    $\ell=74$ ($b=4$) & 
    $\ell=89$ ($b=5$) & 
    $\ell=221$ ($b=10$)
    \\
    \midrule
    Random & 
    $84.86 \pm 0.41$ &
    $89.14 \pm 0.39$ &
    $91.77 \pm 0.32$ &
    $92.54 \pm 0.17$ &
    $94.89 \pm 0.16$
    \\
    \midrule
    Similarity preserving & 
    $87.35 \pm 0.35$ &
    $90.40 \pm 0.09$ &
    $92.66 \pm 0.17$ &
    $92.73 \pm 0.12$ &
    $\mathbf{95.26} \pm 0.09$
    \\
    \bottomrule
    %\noalign{\vskip 1mm}     
  \end{tabular}
  \captionsetup{width=.96\linewidth}
  \caption{Test accuracy (\%) for different codebook widths $\ell$
  (or WLTLS graph widths $b$)
  on \aloi ($K=12,294$). 
  \protect \\
  One-vs-all achieves 95.9\% with $\ell=K=1,000$ binary predictors.
  \protect \\
  Similarity-preserving assignments significantly improve the test performance compared to random assignments.
  % in \secref{sec:exp-exhaustive}.
  %\label{tbl:datasets}
  }
  \vskip .1cm
\end{table}

\subsubsection{\lshtc results of \figref{fig:lshtc1}}

\begin{table}[ht]
  %\vskip -.2cm
  \centering
  \begin{tabular}{llllll}
    \toprule
    Assignment method & 
    $\ell=56$ ($b=2$) & 
    $\ell=79$ ($b=3$) & 
    $\ell=138$ ($b=5$) &
    $\ell=338$ ($b=10$)  &
    $\ell=879$ ($b=20$)
    \\
    \midrule
    Random & 
    $10.17 \pm 0.71$ &
    $13.33 \pm 0.71$ &
    $16.75 \pm 0.69$ &
    $20.50 \pm 0.73$ &
    $21.95 \pm 0.55$
    \\
    \midrule
    Similarity preserving & 
    $13.19 \pm 0.15$ &
    $16.62 \pm 0.25$ &
    $19.22 \pm 0.08$ &
    $22.64 \pm 0.16$ &
    $\mathbf{23.82} \pm 0.19$
    \\
    \bottomrule
    %\noalign{\vskip 1mm}     
  \end{tabular}
  \captionsetup{width=.96\linewidth}
  \caption{Test accuracy (\%) for different codebook widths $\ell$
  (or WLTLS graph widths $b$)
  on \lshtc ($K=12,294$). 
  \protect \\
  One-vs-all achieves 23.3\% with $\ell=K=12,294$ binary predictors.
  \protect \\
  Similarity-preserving assignments beat OVA with only $879$ binary predictors ($\times 14$ less).
  % in \secref{sec:exp-exhaustive}.
  %\label{tbl:datasets}
  }
  \vskip .1cm
\end{table}

\subsubsection{\dmoz results of \figref{fig:lshtc2}}
\begin{table}[ht]
  %\vskip -.2cm
  \centering
  \begin{tabular}{llllll}
    \toprule
    Assignment method & 
    $\ell=62$ ($b=2$) & 
    $\ell=86$ ($b=3$) & 
    $\ell=151$ ($b=5$) & 
    $\ell=351$ ($b=10$) & 
    $\ell=904$ ($b=20$)
    \\
    \midrule
    Random & 
    $11.34 \pm 0.33$ &
    $14.18 \pm 0.11$ &
    $17.39 \pm 0.44$ &
    $21.97 \pm 0.46$ &
    $26.06 \pm 0.23$
    \\
    \midrule
    Similarity preserving & 
    $13.77 \pm 0.12$ &
    $17.24 \pm 0.28$ &
    $20.54 \pm 0.45$ &
    $24.84 \pm 0.31$ &
    $\mathbf{28.31} \pm 0.09$
    \\
    \bottomrule
    %\noalign{\vskip 1mm}     
  \end{tabular}
  \captionsetup{width=.96\linewidth}
  \caption{Test accuracy (\%) for different codebook widths $\ell$
  (or WLTLS graph widths $b$) 
  on \dmoz ($K=27,840$). 
  \protect \\
  One-vs-all achieves 27.88\% with $\ell=K=27,840$ binary predictors.
  \protect \\
  Similarity-preserving assignments beat OVA with only $904$ binary predictors ($\times 30.8$ less).
  % in \secref{sec:exp-exhaustive}.
  %\label{tbl:datasets}
  }
  \vskip .1cm
\end{table}

\subsubsection{\odp results of \figref{fig:odp}}

\begin{table}[ht!]
  %\vskip -.2cm
  \centering
  \begin{tabular}{llll}
    \toprule
    Assignment method & 
    $\ell=72$ ($b=2$) & 
    $\ell=299$ ($b=8$) & 
    $\ell=752$ ($b=15$)
    \\
    \midrule
    Random & 
    $2.71 \pm 0.09$ &
    $9.05 \pm 0.03$ &
    $11.71 \pm 0.47$
    \\
    \midrule
    Similarity preserving & 
    $4.27 \pm 0.02$ &
    $10.45 \pm 0.08$ &
    $\mathbf{12.69} \pm 0.04$
    \\
    \bottomrule
    %\noalign{\vskip 1mm}     
  \end{tabular}
  \captionsetup{width=.96\linewidth}
  \caption{Test accuracy (\%) for different codebook widths $\ell$
  (or WLTLS graph widths $b$) 
  on \odp ($K=104,136$). 
  \protect \\
  Training a one-vs-all model for 104K classes on 423K features is too costly,
  hence we do not report its performance for this dataset.
  }
  %\vskip -.2cm
\end{table}